\documentclass[journal]{IEEEtran}

\usepackage[pdftex]{graphicx}
\ifCLASSINFOpdf
\else
\fi

\usepackage{amsmath}

\usepackage[utf8]{inputenc}
\usepackage{xcolor}
\usepackage{amsthm}

\usepackage{hyperref}

\usepackage[numbers,sort&compress]{natbib}

\usepackage{amsmath}   

\graphicspath{{./}}
 
\DeclareGraphicsExtensions{.png}

\usepackage{booktabs}
\usepackage{textcomp}
\usepackage{tabularx}
\usepackage{arydshln}

\usepackage{float}
\restylefloat{table}

\begin{document}
%
\title{An Evolutionary Game Model for Understanding Fraud in Consumption Taxes}
%
%
%
\author{Manuel Chica, University of Granada, SPAIN\\
        Juan M. Hern\'{a}ndez, University of Las Palmas de Gran Canaria, SPAIN\\
        Casiano Manrique-de-Lara-Peñate, University of Las Palmas de Gran Canaria, SPAIN \\
        Raymond Chiong, The University of Newcastle, AUSTRALIA

\thanks{Corresponding author: Manuel Chica (Email: manuelchica@ugr.es)}
}

\markboth{IEEE Computational Intelligence Magazine}{}

\maketitle

\begin{abstract}

This paper presents a computational evolutionary game model to study and understand fraud dynamics in the consumption tax system. Players are cooperators if they correctly declare their value added tax (VAT), and are defectors otherwise. Each player's payoff is influenced by the amount evaded and the subjective probability of being inspected by tax authorities. Since transactions between companies must be declared by both the buyer and seller, a strategy adopted by one influences the other's payoff. We study the model with a well-mixed population and different scale-free networks. Model parameters were calibrated using real-world data of VAT declarations by businesses registered in the Canary Islands region of Spain. We analyzed several scenarios of audit probabilities for high and low transactions and their prevalence in the population, as well as social rewards and penalties to find the most efficient policy to increase the proportion of cooperators. Two major insights were found. First, increasing the subjective audit probability for low transactions is more efficient than increasing this probability for high transactions. Second, favoring social rewards for cooperators or alternative penalties for defectors can be effective policies, but their success depends on the distribution of the audit probability for low and high transactions. 

\end{abstract}


\section{Introduction}
\label{sec:introduction}

The value added tax (VAT) is the most common consumption tax worldwide. With extensive use since the 1960s, it has reached significant tax revenue capacity. In fact, consumption taxes demonstrated a similar collection capacity to income taxes during the 1990-2010 period~\cite{Prichard2014}. All deliveries of goods and services by companies, professionals, and importers are subject to VAT. The tax base is the value added (value of production minus value of intermediate consumption) generated at each step of the production and distribution process. The tax debt is calculated by applying the tax rate to the value of the goods and services sold, and deducting the VAT attached to intermediate consumption. Companies and professionals declare the VAT passed on to the customers, while deducting the amount borne from purchases from their own suppliers. The resulting statement can be either positive or negative.

This refunding system potentially allows for significant levels of fraud in terms of undervaluation of sales and overvaluation of purchases~\cite{Keen2007}. However, the fact that both buyers and sellers record each transaction offers some self-enforcement capability to the VAT system. Das-Gupta and Gang~\cite{DasGupta03} identified circumstances where the ability of tax administrators to match the sales and purchase invoices strengthens this self-reinforcement capacity. In any case, tracking the relationships between buyers and sellers is clearly of the utmost importance to detect and prevent fraud. For this purpose, tax administration normally requires the declaration of all purchase and sales transactions between pairs that exceed a certain threshold, with an explicit declaration of the counterparts.

In this work, we propose a computational intelligence (CI) model, based on evolutionary game theory~\cite{Nowak92,Shafi17}, to study the VAT fraud dynamics of buyers and sellers in an economic system. The goal of the model is to represent a network of players---either cooperating (correctly declaring VAT) or defecting (incorrectly declaring VAT)---linked by their pairwise transactions. Evolutionary game theory has been applied to models of cooperation such as the well-known prisoner’s dilemma~\cite{Vilone14, Chiong12TEC, Mittal09}, snowdrift games~\cite{Hauert04,Wang06}, or trust dilemmas~\cite{Chica18IEEETEC, Chica19CNSNS}. These game models represent one of the most prominent CI techniques for representing economic markets and designing economic policies~\cite{Dawid08}. The application of evolutionary game models to study tax fraud and evasion is, however, very limited~\cite{Pickhardt14, Antoci14, diMauro19, degiovanni19}, and non-existent when focusing on consumption taxes such as VAT. 

Our CI model is studied in a structured population where players are linked by means of a social network of transactions. Here, players are given two possible strategies: being a cooperator $C$ or being a defector $D$ (i.e., a \textit{free rider}). The model considers the amount of tax accrued on transactions not declared by the free riders, a perceived probability of being inspected by the tax agency, and the corresponding fine when tax evasion is detected. Cooperators, who are players correctly paying their taxes, receive a recognition or social reward; however, they can also have their transactions inspected, with a certain probability, when their transaction records do not match those of their transaction partners in the network.
 
The combination of CI techniques with agent-based modeling (ABM)~\cite{Macal05, Chica17JMR} offers many opportunities for practitioners~\cite{Abbass11}, and our work is a perfect example. Our model represents players of the tax system as agents on nodes of a heterogeneous social network~\cite{Newman06}. The social network follows a power-law distribution, equivalent to the scale-free network topology used in previous studies for promoting cooperation in social dilemmas~\cite{Santos08}. This social network is weighted, with weights of the edges representing values of the tax debt associated with the transactions between two linked nodes. These weights make the model a mixed game~\cite{Amaral16} where players have different payoff matrices depending on the volume of their transactions. The players, through a social evolutionary learning process, can imitate others' strategies by using an evolutionary update rule and a mutation operator to randomly modify their strategy. 

We used real-world data from the Canary Islands tax agency to feed most of the parameters of the model and fit the power-law distribution of the scale-free social network~\cite{Broido19}. After investigating the general dynamics of the model and effects of having well-mixed and structured populations on scale-free networks with different properties, we focused our experiments on determining policies to promote cooperation and reduce the number of players who do not correctly pay their consumption taxes. To achieve this goal, we defined different experimental scenarios that allow us to understand when the best cooperative behaviors occur. These scenarios include policies regarding the shared pressure to increase the perceived probability of being inspected for high and low transactions and how diversity in subjective probabilities affects the levels of cooperation in the population; the impact of modifying the reputational reward for cooperators; and a sensitivity analysis on different inspection fines for defectors or free riders.

In the next section (i.e., Section~\ref{sec:background}), we discuss related work and the motivation of our study. Details of the CI agent-based model are then described in Section~\ref{sec:model}. Section~\ref{sec:data_setup} presents the analysis of real data from a tax agency and setup of the model. The results and model's dynamics are discussed in Section~\ref{sec:results_analysis}. Finally, Section~\ref{sec:conclusions} summarizes the key contributions.

\section{\label{sec:background}Background and related work}


The neoclassical economic model on tax fraud by Allingham and Sandmo~\cite{Allingham1972} is considered one of the cornerstones of the economic analysis of tax evasion. They represent how individual agents decide to evade taxes, while also considering how the government would eventually punish them. However, this model is unable to explain low levels of fraud under low penalty and detection rates. Bordignon~\cite{Bordignon1993} was one of the first to describe this problem and the need to explain tax fraud using explanation other than just selfishness. Subsequent models stress that tax compliance by agents is dependent on how they perceive unfairness in their relations with not only the administration, which is the vertical factor, but also the rest of the agents, which is the horizontal factor~\cite{Bazart2014}. 

When analyzing the horizontal factor, the tax evasion literature tries to identify how the compliance level of an agent affects the compliance level observed by the rest of the agents. Traxler~\cite{Traxler2010} attempted to model different levels of tax evasion within and between groups of agents. He was able to bring issues related to belief management into the discussion, extending the spectrum of policy instruments to the scope of changing individual beliefs, besides the economic incentives. Prichard et al.~\cite{Prichard2014} reflected on the main reasons of the failure of mainstream neoclassical models in their survey. They identified two main lines of research that can address the limitations of the traditional models by including the relevance of behavioral aspects: experiments and ABM.

Experiments, as Alm~\cite{Alm2012} stated, are not without problems, but they overcome the simplicity of theoretical models of individual choice, since they can incorporate many explanatory factors suggested by theory. They also favor the combination of economic theory with other disciplines like psychology, increasing the realism of explanatory factors of tax fraud~\cite{Alm2015b}. 

Bonein~\cite{Bonein2018} has identified different levels of reciprocity between agents. Under “strong reciprocity”, taxpayers would tend to evade more (less) if they observe a more (less) disadvantageous, inequitable behavior by the remaining agents. This completely contradicts the predictions by self-interest models~\cite{Gintis2000}, where agents are only motivated by a future economic benefit. Frey and Torgler~\cite{Frey2007} found strong empirical evidence supporting mutual influence between taxpayers. The investigations by Blaufus et al.~\cite{Blaufus2017} and Calvet and Alm~\cite{Calvet2014} show other perspectives of horizontal factors, such as tax privacy, empathy, and sympathy.

Fedeli and Forte~\cite{Fedeli1999} studied the bargaining problem between agents to decide the level of (under) reporting of sales and purchases along the different steps of the supply chain, combining sales and income tax. Each pair of agents has to decide the amount of trade between themselves and the optimal level of (under) reporting. They concluded that the best policy to control fraud is to increase the inspection probability rather than to increase penalties against fraudulent activities. On the other hand, by acting against one agent, the tax agency may eliminate the incentives for misreporting along the entire chain, thus reducing the enforcement costs for increasing the inspection probability for all agents. Abraham et al.~\cite{Abraham2017} offered another example of modeling based on a non-cooperative game-theoretic model and demonstrated that social norms are more important for explaining behavior in tax fraud than in situations where agents act independently.  

Evolutionary game theory is commonly supported by ABM, a CI simulation paradigm. These CI models start with the conviction that, especially in the field of sales taxes, fraud always involves more than one agent. Boadway et al.~\cite{Boadway2002} presented the payoffs of a group of potential evaders in the form of a prisoner’s dilemma game. With collaboration, they could increase their revenues; but if one of the agents decides not to collaborate, the rest can be negatively affected. Following the results of their model, we can say that a fundamental determinant of tax fraud that requires collaboration between two agents is the capacity of each agent to collaborate in the evasion activity. In the above model, the collaboration capacity depends on the agent’s tolerance of dishonesty. 

There are several advantages of using ABM in the analysis of tax fraud. The ABM approach enables the inclusion of agents with very different response patterns, which allows for a dynamic bottom-up structure that may converge in a global equilibrium that houses an intricate set of individual realities. On the other hand, the agents included in an ABM do not maintain static positions and reactions, but can evolve throughout the simulation, learning from their own situation and those of the other agents. This interaction between agents (i.e., social network) is, therefore, one of the fundamental components of this type of model.

ABM examples in the study of tax fraud can be found in the work of Bloomquist~\cite{Bloomquist2006}, which described this methodology as relatively recent and highlighted the apparent lack of acceptance by the mainstream social scientists. Hokamp et al.~\cite{Hokamp2018} also discussed recent advances in the modeling of tax fraud analysis. The different attitudes of agents may manifest themselves in individual perceptions of both the risk of apprehension and the audit probability~\cite{Chen08}----as in the model presented by Korobow and Axtell~\cite{Korobow2007}. In relation to tax compliance, their model suggested that the outcome depends on how strongly agents are influenced by their neighbor's results. If agents do not significantly value the potential gains of evaders, it is possible to achieve general compliance even if the model realistically incorporates low enforcement measures. Another example with a similar pattern was provided by Hashimzade et al.~\cite{Hashimzade2014}. Combining behavioral economics and social networks, their model reflects how the connection between agents can form a subjective audit probability, which clearly exceeds the objective levels that stem from real levels of fiscal auditing activities. The compliance decisions are modeled to include benefits for the agents for following the social norms of honesty. In a similar fashion, Hashimzade et al.~\cite{Hashimzade2015} described their starting point in terms of attitudes, beliefs and network effects based on two features of empirical analysis on tax fraud. First, there seems to be strong evidence to indicate that individual compliance attitude is clearly affected by social norms. Second, agents do not normally have access to the real audit probability. Therefore, they end up generating their own subjective probability of being subject to inspection and being identified as an evader. A similar basis was used by Hashimzade and Myles~\cite{Hashimzade2017} in their effort to incorporate predictive analysis and behavioral economics in modeling the tax compliance decision.

Previous efforts, such as those in the field of experimental analysis, are very relevant and should continue to be so. Bloomquist~\cite{Bloomquist2011} presented a good example of how the results of experimental economics can be exploited in the development of ABM. New efforts such as those coming from the field of econophysics should be closely observed in order to identify possible complementarities~\cite{PickhardtSeibold2014}. It has already been mentioned that network tools are an essential part of ABM; therefore, delving into how the network structure influences the results of these models is an important avenue of study. The first step in this direction came from Andrei et al.~\cite{Andrei2014}, who explored the impact of different network structures on the levels of tax evasion in an agent-based model. 

Reality is complex and models seeking to describe it must be able to reflect this complexity. Calibrating the models to real data is not an easy task when dealing with ABM and tax fraud. However, linking the main elements of an agent-based model to real data or observed behavior of the relevant agents must always be an end to pursue. Given that the data necessary for this type of analysis is inexorably linked to individual data protection rights, collaboration between researchers and tax administration is a must. This is especially the case if we consider increasing our knowledge of taxpayer compliance behavior, as indicated by Bloomquist~\cite{Bloomquist2012}.

\section{\label{sec:model}Model description}   

\subsection{\label{sec:model_desc} Game Strategies and Payoff Matrix}

Players of the game are a finite set of $Z$ agents (companies) occupying the nodes of a social network, with edges denoting economic transactions between them. The network is undirected but weighted: a weight $d_{ij}$ means the accrued tax to be declared and paid in relation with all cumulative transactions between both players $i$ and $j$. Under a correct behavior, this value $d_{ij}$ is the consumption tax involved in the transactions between both companies $i$ and $j$. We differentiate between two quantities to be paid, high ($d_H$) and low ($d_L$), based on high and low transaction values, respectively. Therefore, $d_{ij}=\{d_L,d_H\}, \forall \langle i,j \rangle$ being an edge of the social network. In this sense, the game can be called a mixed game~\cite{Wardil13, Amaral16} where two players play a different game depending on the type of transaction, given by parameter $d_{ij}$.

Each player $i$ chooses a strategy $s$ from two possibilities at every time step ($s(i)=\{C,D\}$): being a cooperator or tax payer ($C$), or being a defector or tax evader ($D$). When being a tax evader, the player does not pay a fraction of the transaction value $d_{ij}$, saving this cost as a personal benefit (\textit{free rider}). We model this fraudulent fraction by parameter $\alpha \in [0,1]$, which also measures the difficulty of the social dilemma. Higher values of $\alpha$ correspond to higher economical benefits for free riders when declaring transactions.

In order to set the fitness of a player $i$ for a specific time step $t$ ($f_i^t$), the player accumulates all the payoffs $w^t_{ij}$ from the pairwise interactions over all its direct contacts in the social network (i.e., neighborhood in the network): $f_i^t = \frac{1}{\langle k \rangle_i} \sum^{\langle k \rangle_i}_{j=1} w^t_{ij}$, where $\langle k \rangle_i$ is the degree of player $i$. The payoff $w_{ij}$ of focal agent $i$ with respect to neighbor $j$ is obtained by considering their strategies in the previous time step and the specific payoff matrix played depending on the type of transaction (either low or high). 

Table~\ref{tab:utility_matrix} shows the payoff matrix defining the mixed game. Parameter $R$ is the social and reputational reward for a player when acting in accordance to its tax duties. $\Gamma$ is the inspection cost a company should pay when a tax agency examines the company and its documentation regardless of its own behavior (i.e., playing strategy). $\phi$ is the fine multiplier a company must pay when the tax agency audits the company and discovers fraudulent behavior.

\begin{table*}[!ht]
\caption{Payoffs for the mixed tax fraud game.} \label{tab:utility_matrix}
\centering
\begin{tabular}{l | c @{\hskip 0.5in} c }   

 & $C$ & $D$ \\
\toprule

$C$ &  $R$           &     $R - \Theta   \Gamma$  \\
 $D$ &  $\alpha  d -  \Theta  [\Gamma + \phi  \alpha  d] $   &     $\alpha  d - \Theta' [\Gamma + \phi \alpha  d]  $    \\ 
\end{tabular}
\end{table*}

\begin{table*}[ht]
\centering
\begin{tabular}{cl}
\toprule

\textbf{NAME} & \textbf{DESCRIPTION} \\
\midrule

$R$ &  REPUTATIONAL AND SOCIAL REWARD FOR CORRECTLY PAYING TAXES \\
$\Gamma$ & INSPECTION COST \\
$d = \{d_H, d_L\}$ &  AMOUNT OF TAX DEBT TO BE PAID FOR THE INVOLVED TRANSACTIONS \\
$\alpha \in [0,1]$ &  RATIO OF UNPAID TAX DEBT BY DEFECTORS  \\
$\phi$ & FINE MULTIPLIER TO BE PAID BY A DEFECTOR UPON INSPECTION \\ 
$\Theta = \{\Theta_H, \Theta_L\}$ & TAXPAYER'S SUBJECTIVE PROBABILITY OF BEING AUDITED WHEN ONE PLAYER DEFECTS \\
$\Theta_H = \Theta(\alpha d_H)$ & SUBJECTIVE PROBABILITY FOR HIGH TRANSACTIONS WHEN ONE PLAYER DEFECTS \\
$\Theta_L = \Theta(\alpha d_L)$ & SUBJECTIVE PROBABILITY FOR LOW TRANSACTIONS WHEN ONE PLAYER DEFECTS  \\
$\Theta^\prime = \{\Theta^\prime_H, \Theta^\prime_L\}$ & TAXPAYER'S SUBJECTIVE PROBABILITY OF BEING AUDITED WHEN BOTH PLAYERS DEFECT \\
$\Theta'_H = \Theta(2 \alpha d_H)$ & SUBJECTIVE PROBABILITY FOR HIGH TRANSACTIONS WHEN BOTH PLAYERS DEFECT  \\
$\Theta'_L = \Theta(2 \alpha d_L)$ & SUBJECTIVE PROBABILITY FOR LOW TRANSACTIONS WHEN BOTH PLAYERS DEFECT  \\

\bottomrule

\end{tabular}
\caption{List of parameters of the tax fraud evolutionary game. $\Theta(\cdot)$ is a linear function.}. 
\label{tab:parameters}

\end{table*}

$\Theta()$ is a linear probability function to define a player's perception of how probable a tax audit is. This probability is subjective, even if all the players have the same perception of this probability (we can say there is a ``shared collective perception''), and depends on the taxpayer's subjective probabilities about being audited, following previous studies such as that of Hashimzade et al.~\cite{Hashimzade2014}. The function $\Theta()$ depends on the difference in the amount declared by the players, and is only applied when the tax agency discovers a transaction mismatch for a pair of players. The probability function is built from two values ($2\alpha d_L$ and $2\alpha d_H$), which set the probability of being inspected for both low and high transactions. For clarity, we have defined $\Theta_H$, $\Theta_L$, $\Theta'_H$, and $\Theta'_L$ as the values of the probability function based on its arguments. Table~\ref{tab:parameters} shows a summary of all the model parameters.

The payoff matrix in Table~\ref{tab:utility_matrix} includes different two-strategy games depending on the values of the parameters. These games determine the decision of every player to either cooperate or defect, and therefore influence the final outcome of the evolutionary game. To characterize the included games we consider a general payoff matrix:

\begin{center}
\begin{tabular}{l | c  c }  

 & $C$ & $D$ \\
\toprule

$C$ &  $R$           &     $S$  \\
$D$ &  $T $   &     $P $    \\

\end{tabular}
\end{center}
\noindent where $S$, $T$, and $P$ represent the expressions of Table~\ref{tab:utility_matrix}. The definition of the payoff matrix in Table~\ref{tab:utility_matrix} satisfies $R>S$ and $T>P$, facilitating the level of cooperation in the game. According to Allen and Nowak~\cite{Allen2015}, a social dilemma occurs when $R>P$ (mutual cooperation benefits both players) and at least one of the following conditions is met to favor the adoption of defection: $(D_1): T>R$, $(D_2): P>S$, or $(D_3): T>S$. Values of the parameters in Table~\ref{tab:parameters} determine which of these conditions are satisfied.

\begin{figure}
\centering
\includegraphics[scale=0.55]{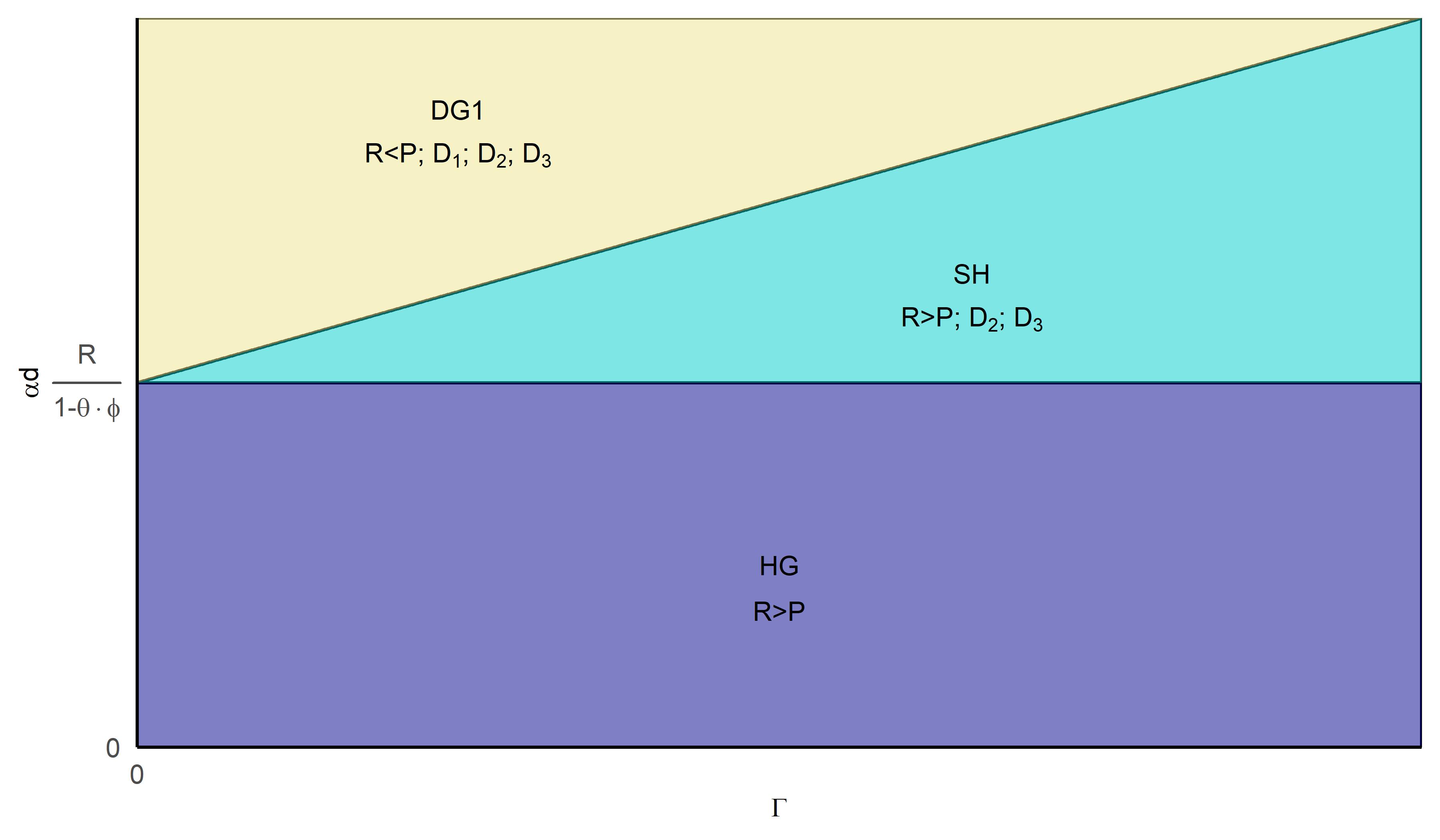}
\caption{Different games for the payoff matrix in Table~\ref{tab:utility_matrix} according to parameter values $\Gamma$ and $\alpha d$. The subjective audit probability is a constant function $\Theta(\cdot)=\Theta$. Here, $SH$ is the Stag Hunt game; $HG$ is a Harmony Game, and $DG1$ represents a defection game with conditions $D_1$, $D_2$ and $D_3$}. 
\label{games_gamma}
\end{figure}

Figure~\ref{games_gamma} shows possible games according to $\alpha d$ and inspection cost $\Gamma$ values; and assuming a constant audit probability $\Theta(\cdot)=\Theta$. We see three regions in the figure depending on the parameters' values: there is no social dilemma in two of them and the third is associated with a classical game. Cooperation prevails if the non-declared amount $\alpha d$ is below $R/(1-\Theta\phi)$, whereas defection is the preferred option if $\alpha d$ is high enough. A Stag Hunt game appears for intermediate values of $\alpha d$, where conditions for cooperation $R>P$ and temptation to defect ($D_2$ and $D_3$) coexist.

\begin{figure}
\centering
\includegraphics[scale=0.55]{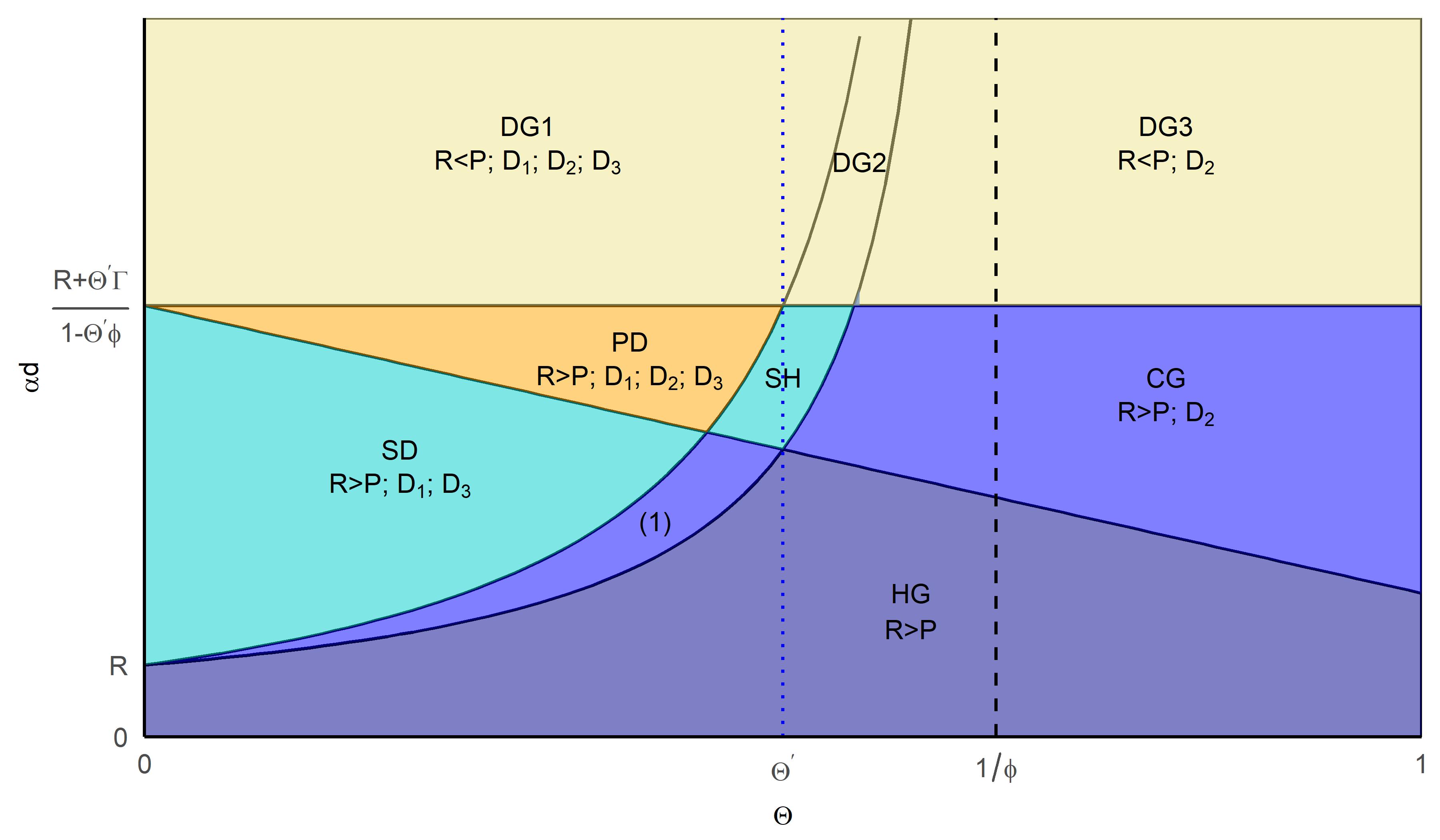}
\caption{Different games for the payoff matrix in Table~\ref{tab:utility_matrix} according to parameter values $\Theta$ and $\alpha d$. $PD$ is the Prisoner's Dilemma game; $SD$ is the Snowdrift game; $CG$ is a coordination game; $DG2$ represents a defection game with condition $D_2$ and $D_3$; $DG3$ is a defection game with condition $D_2$; and $(1)$ is a game with $R>P$ and $D2$.}
\label{games_theta}
\end{figure}

In case of a non-constant audit probability $\Theta(\cdot)$, the outcome is more complicated. The payoff matrix includes two possible values, $\Theta(\alpha d)=\Theta$ and $\Theta(2 \alpha d)=\Theta^\prime$. Figure \ref{games_theta} shows the set of games for different values of $\Theta$ and $\alpha d$, assuming a fixed $\Theta^\prime$. For large values of $\alpha d$ (above the horizontal line defined by $(r+\Theta^\prime \Gamma)/(1-\Theta^\prime \phi)$ ), there is no social dilemma as mutual defection is always preferred over mutual cooperation ($P>R$). Below the horizontal line, multiple games arise. In general, cooperation is expected when $\alpha d$ is low and $\Theta$ is high. More specifically, when the audit probability is a decreasing function ($\Theta>\Theta^\prime$), the most expected games are those favoring cooperation (coordination and harmony games). However, when the audit probability is increasing ($\Theta<\Theta^\prime$), there is a significant region where games such as the prisoner's dilemma or snowdrift prevail and defection is the expected outcome. 

\subsection{\label{sec:update_rules} Evolutionary Update Rule}

The players can change their strategies $s(i), \forall i \in Z$ during the whole discrete-event simulation. These changes in strategies come from two evolutionary mechanisms. First, a player $i$ can imitate others in the population (generally, their direct contacts in the social network). Second, players can also change their strategies by adopting a strategy at random, following a random mutation mechanism with probability $\mu$. The mutation operator does not take into account if the new strategy was beneficial in the past in terms of the fitness values of the players. However, the social imitation update rule is a social learning process of the players in the game~\cite{Nowak10}. Social imitation update rules consider the fitness of direct neighbors on a network in the previous steps to make their decision, either in a deterministic or a probabilistic way. In our model, we use the Fermi function as the social imitation update rule. 

This Fermi rule is one of the most well-known imitation processes~\cite{Gois19, Vasconcelos14} and is applied synchronously: for each step $t$, a focal agent $i$ compares its fitness value in the previous step $t-1$, $f_i^{t-1}$, with one of its direct neighbors in the social network, $j$, also in $t-1$: $f_j^{t-1}$. Therefore, the Fermi rule is a stochastic pairwise comparison rule, where players can also make mistakes during the imitation process (i.e., a player can imitate a neighbor with a worse fitness). Mathematically, agent $i$ with strategy $X$ adopts strategy $Y$ of agent $j$ (a randomly selected direct contact of $i$) with a probability given by Equation~\ref{eq:prop_imitation}, where $\beta$ is the intensity of selection parameter, set to $0.5$ in the model.

\begin{equation}
 prob_{ij}^t = \frac{1}{1 + e^{- \beta(f^{t-1}_j - f^{t-1}_i)}}. 
\label{eq:prop_imitation}
\end{equation} 
\noindent

\section{\label{sec:data_setup}Data analysis and parameters of the model}

\subsection{\label{sec:data_desc}Data Description}



Most of the parameters in the model were set using real data. The real data used for our study includes VAT declarations by businesses registered in the Canary Islands (Spain) in 2002 about transactions, with persons or firms, exceeding 3,005.06 euros. The anonymized data was accessible only within the tax administration under confidentiality agreement of not revealing any information that could be used to identify either the buyers or sellers, and under the commitment that the data was not to be passed on to third parties in any format. 

Every taxpayer must independently declare purchases and sales. Therefore, all transactions between two firms should be declared twice---once each by the seller and the buyer. The original information is split into two files: one with the total sales in the economy and the other with total purchases. A database was built merging both files and taking the transactions (sales and purchases in 2002) declared by the same firms in the two files into account. Transactions without a counterpart were eliminated, since in some cases (e.g., individual buyers) a counterpart is not obliged to declare. Transactions where both counterparts match were also removed. Finally, the number of firms in the database is $N=32,886$ including $E=197,791$ operations.

\subsection{\label{sec:exp_sn}Generation of the Real Social Network of Transactions}

\begin{figure}
\centering
\includegraphics[scale=0.55]{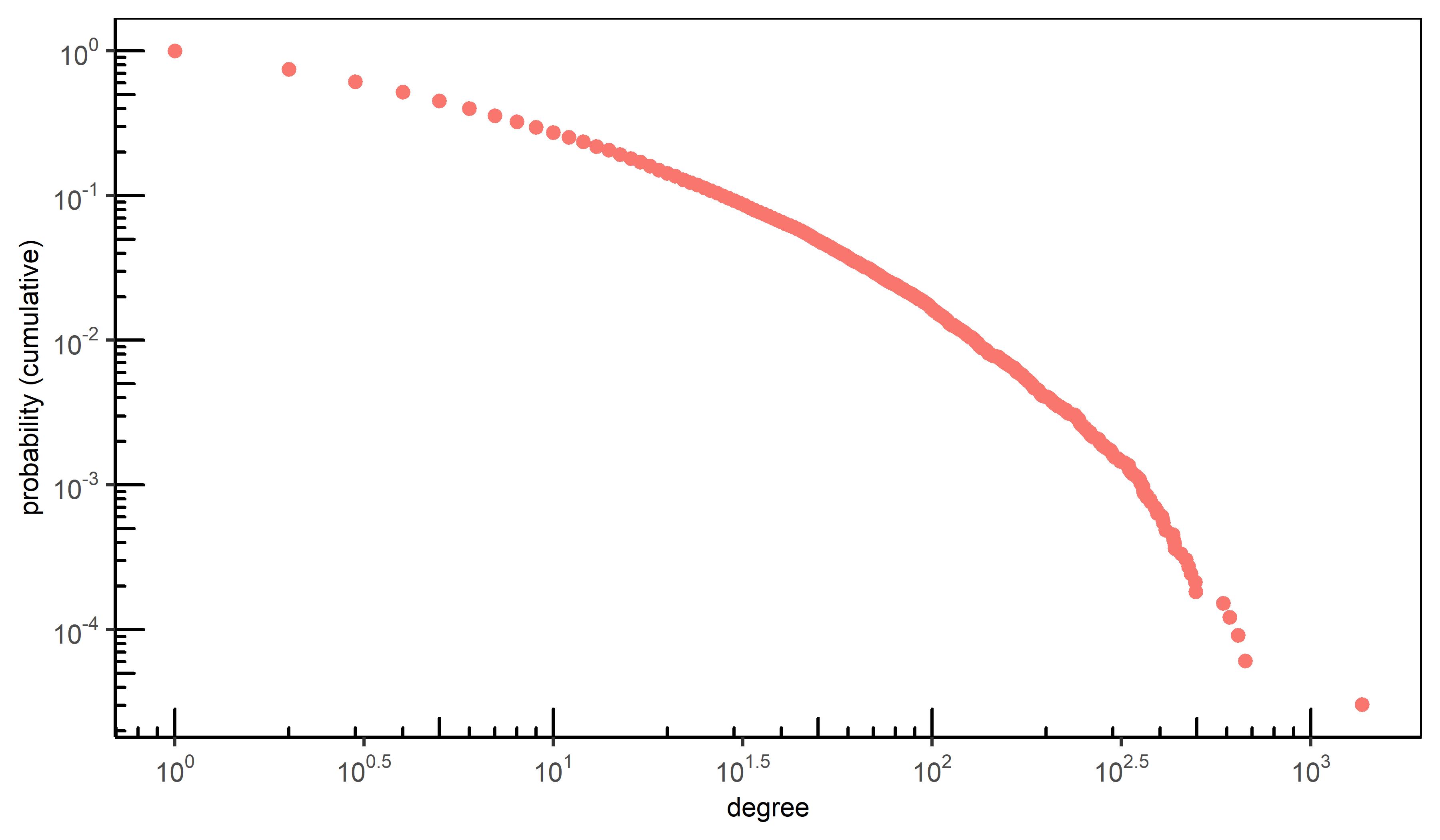}
\caption{Degree distribution of the VAT declaration network.}
\label{degree_dist}
\end{figure}

In order to examine the structure of VAT declarations, we set up a network of firms, where two firms are linked when they declare transactions between them. The network is undirected, since we do not differentiate whether the firm acts as a buyer or a seller, although the amount declared by the seller is included as a link weight. Specifically, we use a scale-free network starting with the degree distribution of the VAT declaration network (Figure \ref{degree_dist}), which is a long-tailed distribution. Following the methodology in Clauset et al.~\cite{Clauset2009a}, the network fits to a power law $k^{-\gamma}$ with $x_{min}=88$ and $\gamma=3.04$. According to the taxonomy in Broido and Clauset~\cite{Broido19}, the VAT declaration network is a weak scale-free network, since its power law cannot be rejected and it includes more than 50 nodes. For the model, we have built a scale-free network of 10,000 nodes, with the same exponent $\gamma$ and $x_{min}$. 

\subsection{\label{sec:exp_quantities}Feeding Transaction Parameters from Data}

\begin{figure}
    \centering
    \includegraphics[scale=0.55]{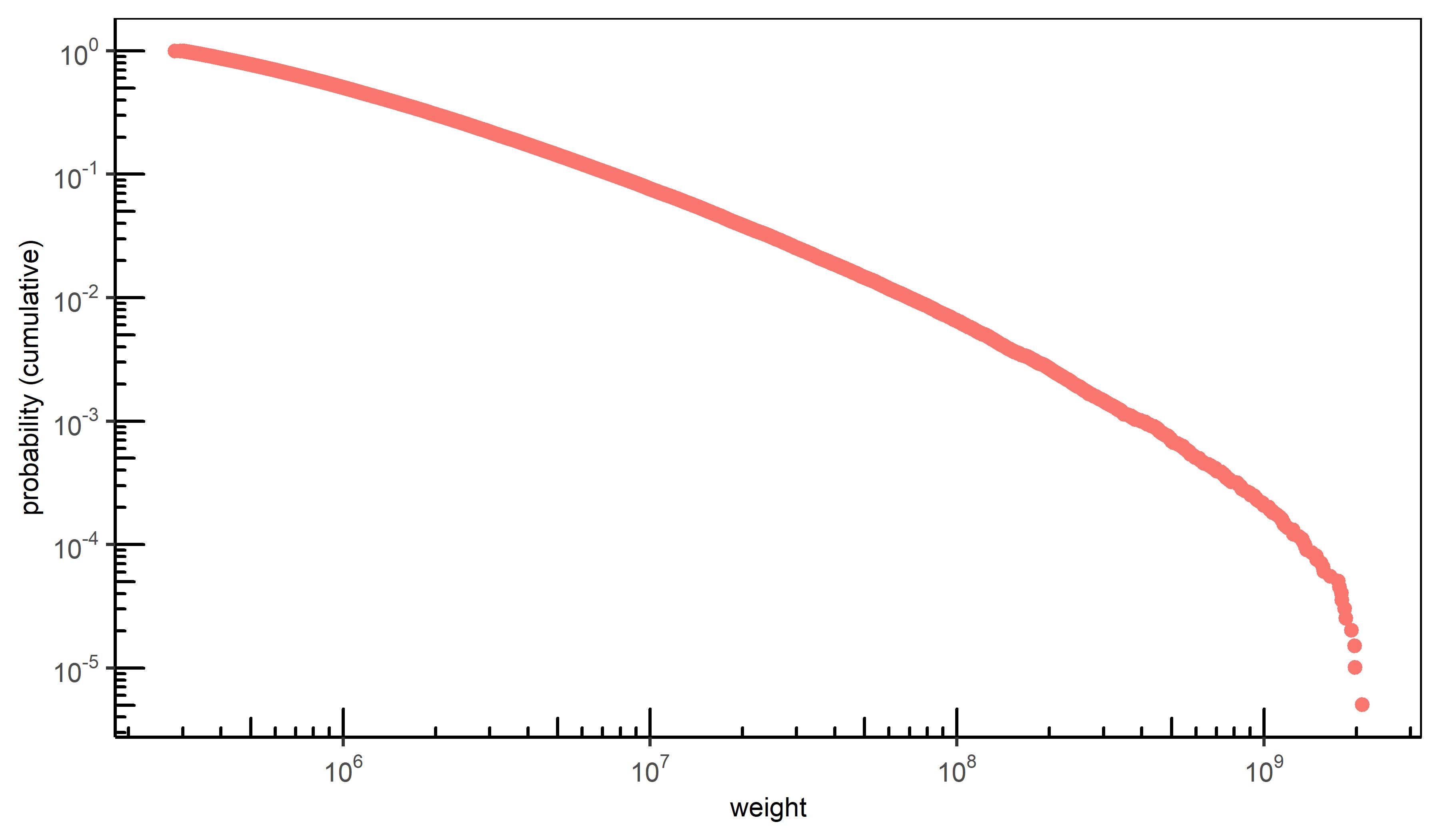}
    \caption{Edge-weight distribution in the real VAT declaration network.}
    \label{edge_w_dist}
\end{figure}

The values and distribution of the quantities to be paid $\{d_L,d_H\}$ were set from the empirical data. Figure~\ref{edge_w_dist} shows the edge-weight distribution by considering sellers' declaration (note that we obtained a similar distribution when we looked at buyers' declaration). More than 95\% of the transaction amounts declared are under 1 million euros, whereas 0.01\% are over 100 million euros. On the other hand, according to official records, the percentage of large firms (those with more than 20 employees) in the Canary Islands is about 2\%~\cite{PYME19}. Therefore, we set the probability of an edge to have a high transaction to $0.02$ (i.e., $prob_{d_H}=0.02$). Additionally, we obtained a ratio $r=45.7589$ between the average values of the total volume of transactions of small and big firms (those transactions below and above quantile 2\% in the edge-weight distribution, respectively). We assume that the amount to be paid is a constant fraction of the transaction value. Therefore, we set the value $d_{ij}=\{d_L,d_H\}$, $\forall \langle i,j \rangle \in E$, to a fixed value of $d_L=10$ and $d_H=rd_L=457.59$. Edges' weights $d_{ij}$ are randomly initialized, characterized by $prob_{d_H}=0.02$. Note that weights $d_{ij}$ do not change over time. They are static and therefore, the same payoff matrix of the mixed game is used for every pair of players $i$ and $j$ in this study.

We define the ratio of divergence in the tax declarations between seller $i$ and buyer $j$ ($\alpha_{i \leftrightarrow j}$) as the percentage of VAT declaration mismatch between the tax declared by seller $i$ and buyer $j$. We only take those mismatches benefiting the sellers (this is the case where the seller declares less amount than the buyer). The ratio is calculated as follows: Given $d_{ij}$ as the amount that firms $i$ and $j$ need to declare, let $\alpha_{i \rightarrow j}$ be the percentage of this amount that seller $i$ does not declare and $\alpha_{j \rightarrow i}$ the percentage that buyer $j$ declares in excess. Then, the ratio of undeclared accrued tax between $i$ and $j$ is: 

\begin{align*}
\alpha_{i \leftrightarrow j}=\frac{(1+\alpha_{j \rightarrow i})d_{ij}-(1-\alpha_{i \rightarrow j})d_{ij}}{(1+\alpha_{j \rightarrow i})d_{ij}+(1-\alpha_{i \rightarrow j})d_{ij}}=
\\\frac{\alpha_{i \rightarrow j}+\alpha_{j \rightarrow i}}{2+\alpha_{j \rightarrow i}-\alpha_{i \rightarrow j}}, \forall \langle i,j \rangle \in E
\end{align*}

This ratio is between 0 (when both counterparts declare exactly the right amount, $\alpha_{i \rightarrow j}=\alpha_{j \rightarrow i}=0$) and 1 (when the seller does not declare any amount). Note that when the fraction of the amount that is incorrectly declared is constant and identical for any firm and transaction, we have $\alpha_{i \rightarrow j}=\alpha_{j \rightarrow i}=\alpha$, with $\alpha$ being the ratio of unpaid quantity for defectors (Table \ref{tab:parameters}). We also have $\alpha_{i \leftrightarrow j}=\alpha, \; \forall \langle i,j \rangle \in E.$ Then, the assigned values for parameter $\alpha$ in the model would be calibrated by the $\alpha_{i \leftrightarrow j}$ values in the real data. 

\begin{figure}
    \centering
    \includegraphics[scale=0.55]{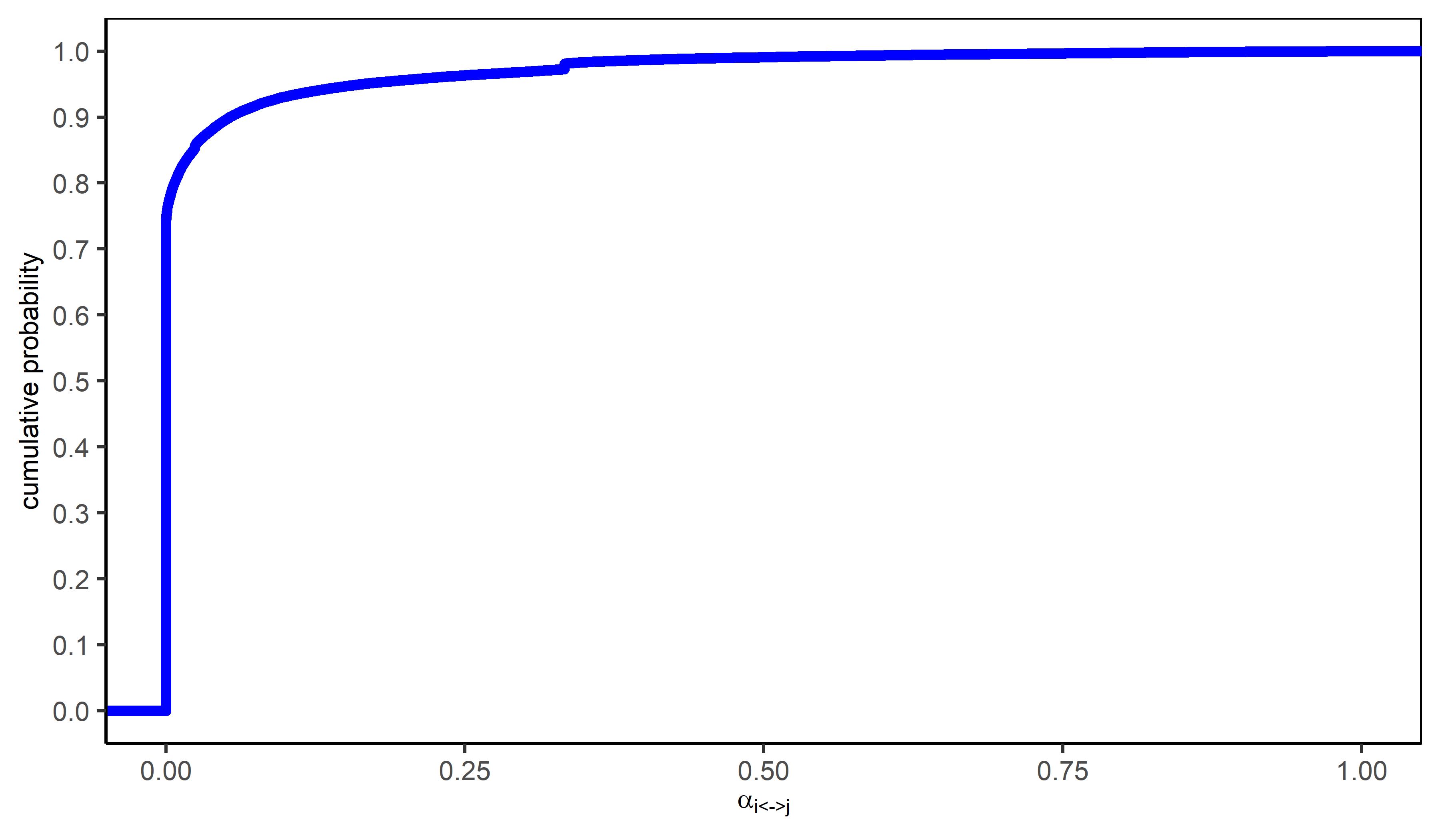}
    \caption{Cumulative distribution of the ratio of undeclared transactions in the VAT declaration network.}
    \label{fig:alpha_ij}
\end{figure}

Figure \ref{fig:alpha_ij} represents the cumulative distribution of the ratio $\alpha_{i \leftrightarrow j}$ in the VAT declaration network. As can be observed, the ratio of undeclared transactions is almost zero for around 75\% of operations and below 0.5 for 99.06\% of them. Given this, in our simulations, we consider the range of realistic values for parameter $\alpha$ between $0$ and $0.5$.

\subsection{\label{sec:exp_setup}Model Setup}

We set the model up for $50$ Monte-Carlo runs, with 1,000 time-steps in each run, thereby ensuring that all the realizations reach a stationary stable state (as we can see in Section~\ref{sec:general_dynamics}). The simulation results were obtained by averaging the last 25\% of the simulation time-steps in the independent Monte-Carlo runs. Source code and data files are available at \textit{}{http://bitbucket.org/mchserrano/evolutionary-game-tax-fraud}.

We set the remaining parameters---when not explicitly specified---as follows: inspection cost $\Gamma=1$ and reputational reward for cooperators $R=1$. Fine value $\phi$ was set to $1.5$ (50\% fine plus the undeclared quantity) as per previous models~\cite{Hashimzade2014}. The values used to generate the linear audit probability function were set to $\Theta'_H = \Theta'_L = 0.5$, when the analysis was not focused on these subjective probabilities (see Section~\ref{sec:results_probs_high_low}). The mutation probability of the evolutionary dynamics was always set to $\mu=0.01$.

\section{\label{sec:results_analysis}Analysis of the results}

\subsection{\label{sec:general_dynamics}General Dynamics of the Model} 


\begin{figure}
\centering
\includegraphics[width=0.48\textwidth]{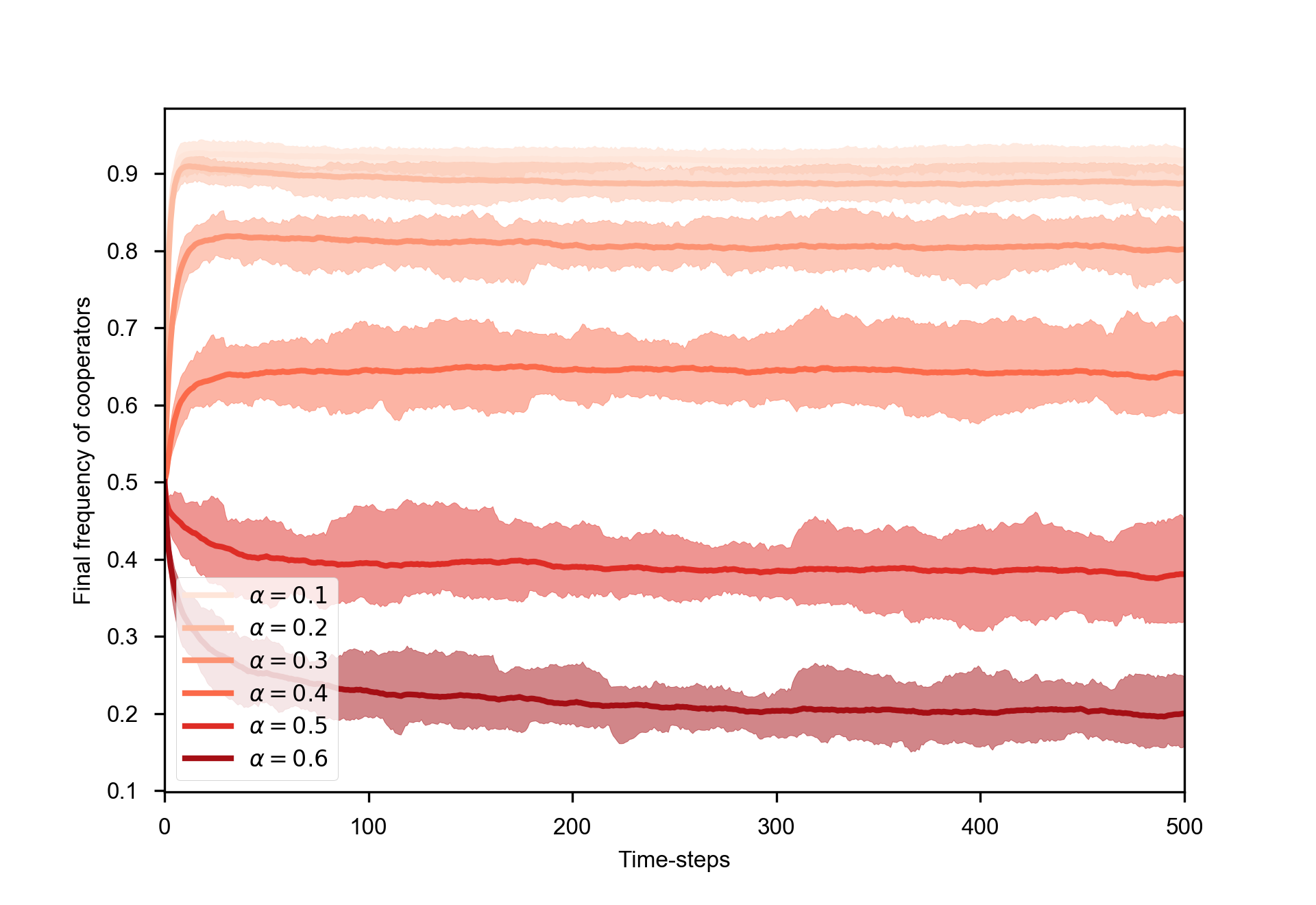}
\includegraphics[width=0.48\textwidth]{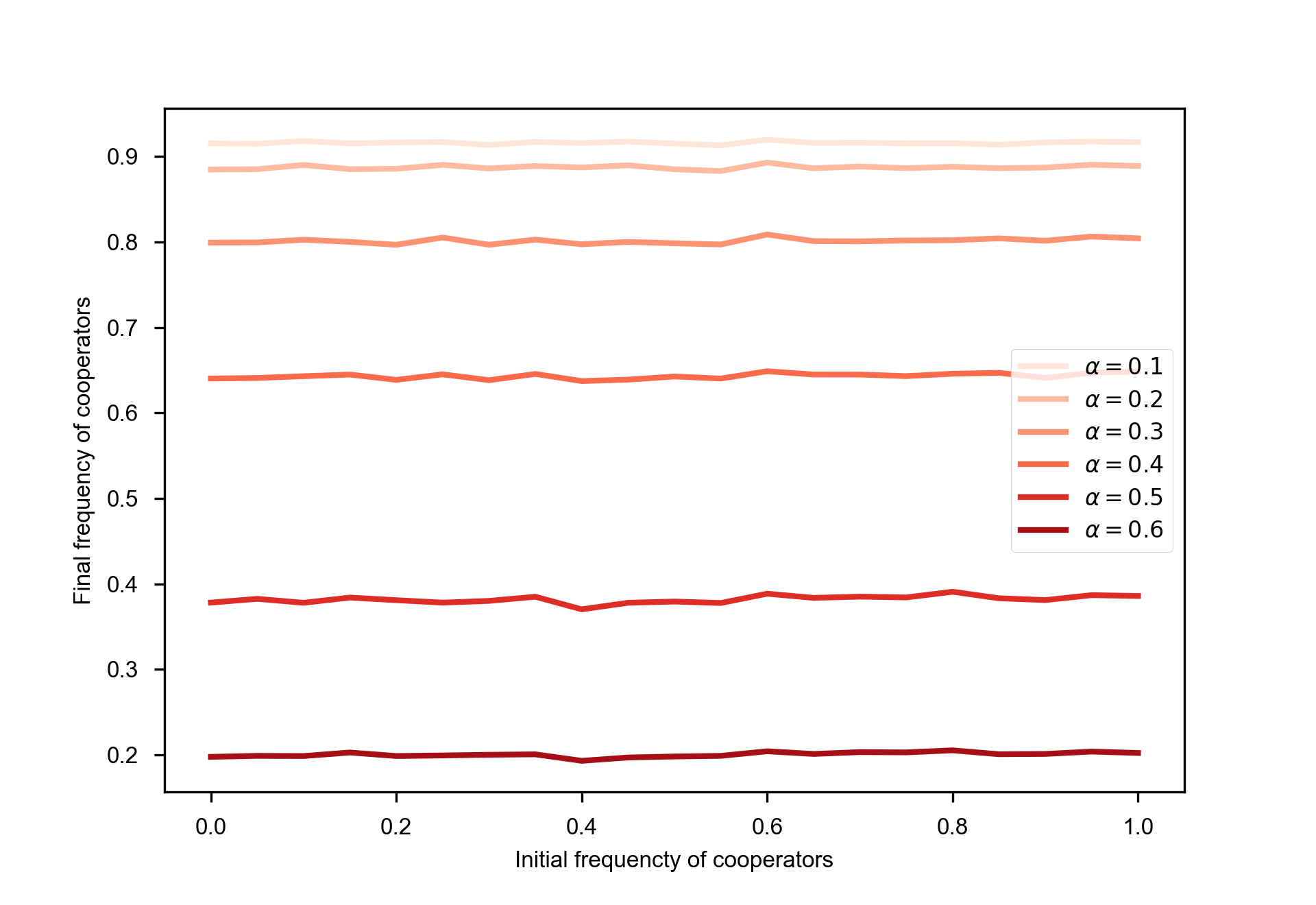}

\caption{The upper plot shows the cooperators' evolution for different $\alpha$ values, from $0.1$ to $0.6$, with an initial cooperators frequency of $0.5$. The lower plot shows that the initial frequency of strategies in the population is not relevant for the final state of the model.}
\label{fig:time_evolution_SA_pop}
\end{figure}

We first show the dynamics of the model for the base parameters. The upper plot in Figure~\ref{fig:time_evolution_SA_pop} shows the evolution of the model over 1,000 time-steps for different values of $\alpha$. The plot also shows the max-min range of the simulations for the $50$ Monte-Carlo realizations. The stationary state is quickly reached and even $500$ time-steps are sufficient in this case. Additionally, the max-min ranges in the plot highlight that the deviation of the model is low. The lower plot in Figure~\ref{fig:time_evolution_SA_pop} shows how the model dynamics is independent of the initial strategy settings of the population. By enabling a player to randomly change their strategy using the mutation operator, we also eliminate differences in the outputs in case of extreme conditions (e.g., the initial frequency of cooperators is either $0$ or $1$). Therefore, we fixed the initial frequency of cooperators to $0.5$ for the rest of the analysis.

\begin{figure}
\centering
\includegraphics[width=0.48\textwidth]{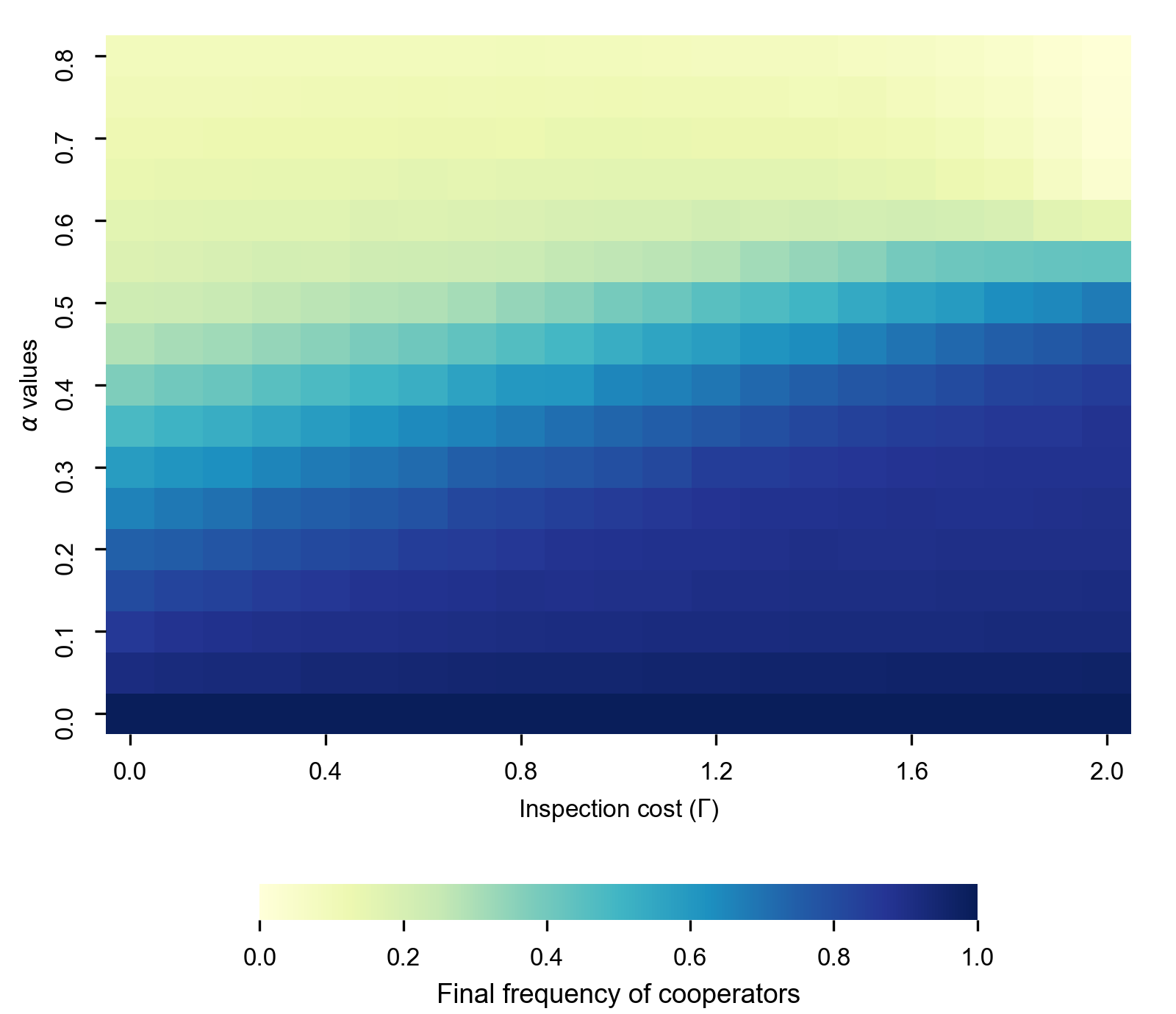}

\caption{Sensitivity analysis on $\alpha$ and $\Gamma$. $\alpha$ controls the difficulty of the game, directly affecting the cooperation level. $\Gamma$ is the inspection cost when a player is inspected due to mismatch declaration. We see how inspection cost $\Gamma$ is only significant for $\alpha$ values approximately between $0.2$ and $0.6$. If $\alpha$ is either lower or higher, $\Gamma$ has no impact on the cooperation level.}
\label{fig:heatmap_SA_alphas_inspCost}
\end{figure}


We also analyze the impact of changing the $\alpha$ values and inspection cost $\Gamma$ by running a sensitivity analysis. Results can be observed in the heatmaps of Figure~\ref{fig:heatmap_SA_alphas_inspCost}.
This graph is in agreement with the set of possible games analyzed in Figure~\ref{games_gamma}. These results show how the model is sensitive to the values of $\alpha$, which smoothly regulates how the population converges to cooperation or defection and therefore, the difficulty of the game. The impact of inspection cost $\Gamma$ is less significant in the model dynamics when $\alpha$ has either high or low values. In fact, when $\alpha$ values are high (or low) and therefore, cooperation (or defection) is restricted, the consequence of changing $\Gamma$ values is minimal. The role of the inspection cost is significant only when the game has an intermediate level of difficulty and, as expected, increasing inspection cost $\Gamma$ promotes higher cooperation. This means that relying on the inspection cost to promote tax compliance is more worthy when the population has a mixture of cooperators and defectors.

\subsection{\label{sec:SF_study}Scale-free Networks Analysis}

\begin{table}[ht]
\small
\centering
\caption{Main features of the social network topologies}
\label{tab:sn_metrics}
\begin{tabular}{cccc}
\toprule
        \textbf{\begin{tabular}[c]{@{}c@{}} \\ \end{tabular}}  \begin{tabular}[c]{@{}c@{}}\textbf{NETWORK}\end{tabular} & \begin{tabular}[c]{@{}c@{}}\textbf{AVERAGE}\\ \textbf{DEGREE}\end{tabular} & \begin{tabular}[c]{@{}c@{}} \textbf{CC} \end{tabular} & \textbf{D} 
         \\    
\midrule
    REAL NETWORK FROM DATA	   & 		1.9952   &  0  & 	111 \\ 
    
    		             & & & \\ 			
    SF (BA WITH $m=2$)	   & 		3.0090	   & 0.0035   & 	12\\ 
    SF (BA WITH $m=4$)	   & 		5.0070   & 	0.0044   & 	9\\
    SF (BA WITH $m=6$)	   & 		7.0100   & 	0.0065   & 	9\\ 
    SF (BA WITH $m=8$)	   & 		9.0220   & 	0.0082   & 	8\\ 
    		             & & & \\  			
    ASS. SF  ($p=0.5, m=2$)    & 			2.9970   & 	0.0029   & 	20\\
    ASS. SF  ($p=1, m=2$) 	   & 		2.9850   & 	0.0120   & 	283\\ 
    DISS. SF  ($p=0.5, m=2$) 	   & 		2.9935	   & 0.0016   & 	18\\ 
    DISS. SF  ($p=1, m=2$) 	   & 		2.9860   & 	0.0002	   & 15\\ 
    		             & & & \\ 			
    ASS. SF  ($p=0.5, m=8$) 	   & 		9.0040   & 	0.0069   & 	11\\
    ASS. SF  ($p=1, m=8$) 	   & 	9.0460   & 	0.0325	   & 53\\ 
    DISS. SF  ($p=0.5, m=8$) 	   & 		9.0120	   & 0.0097   & 	8\\
    DISS. SF  ($p=1, m=8$) 	   & 	8.9820	   & 0.0005   & 	19\\ 
    			
\bottomrule
\end{tabular}
\end{table}

Here we compare the dynamics of different scale-free networks with respect to a well-mixed population. Apart from the fitted real scale-free network, we considered four networks generated by the Barabasi-Albert (BA) algorithm~\cite{Barabasi99} for different values of parameter $m$, controlling the average degree of the networks. Additionally, we used the Xulvi-Brunet-Sokolov algorithm~\cite{Xulvi04} to obtain assortative and disassortative scale-free networks. The assortativity property of networks denotes the preferences of highly connected nodes to be connected with other highly connected nodes~\cite{Newman06}. On the other hand, the disassortativity property denotes the preference of highly connected nodes to connect with less connected nodes. Parameter $p$ of the Xulvi-Brunet-Sokolov algorithm is used to control the degree of assortativity and disassortativity of existing scale-free networks. In our case, we applied the algorithm to the most and least dense networks generated by the BA algorithm (i.e., $m=2$ and $m=8$). Thanks to these network generation algorithms, we employed 12 networks with diverse clustering coefficient (CC), diameter (D), and density. Table~\ref{tab:sn_metrics} lists the features of the networks.
   
\begin{figure}
\centering
\includegraphics[width=0.48\textwidth]{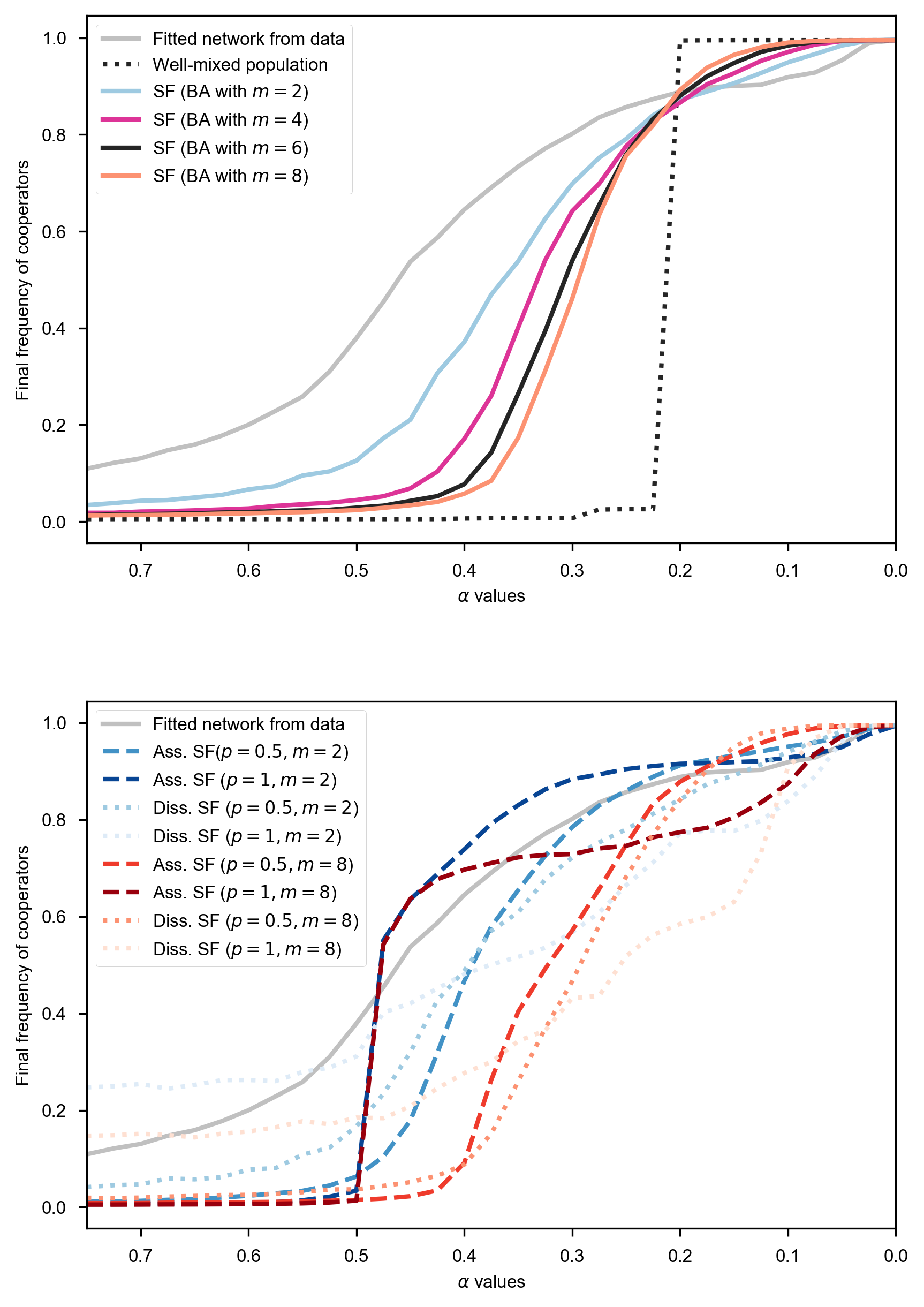}

\caption{The upper plot shows the comparison among the well-mixed population, a network obtained from real data, and scale-free networks generated by the BA algorithm for different $\alpha$ values. The lower plot shows the comparison among different levels of assortativity and disassortativity in the scale-free networks. Density and assortativity impact the level of cooperation depending on the $\alpha$ values. A well-mixed population can only achieve cooperation when the game is trivial ($\alpha \leq 0.2$).}
\label{fig:different_SFs}
\end{figure}

Figure~\ref{fig:different_SFs} shows the dynamics of the model with the networks and a well-mixed population. We can see in the upper plot of Figure~\ref{fig:different_SFs} that cooperation is non-existent with a well-mixed population, except when $\alpha$ is lower than or equal to $0.2$. Note that the expected level of coexistence in the well-mixed population can be analytically derived under some settings of the tax fraud game. In the upper plot of the figure, we can also see that the trends of the BA scale-free networks are similar. Networks with lower density ($m=2$ and fitted network from data) are able to better promote cooperation when $\alpha$ is increasing (the game is harder). When the game is easy (low $\alpha$ values), higher density is better for achieving total cooperation because it increases the speed of diffusion through the network. These results are in line with the well-mixed population output, which jumps from total defection to total cooperation when the game is easy. This abrupt shift in the model results is in agreement with the observation in Figure 1, where we have two extreme cases (defection and harmony games) as the most prevalent games for the parameter values. 

The lower plot of Figure~\ref{fig:different_SFs} shows the dynamics of the model with the above assortative and disassortative topologies and a well-mixed population. As observed with low density networks, disassortativity favors cooperation when the game is hard (high values of $\alpha$). Assortativity plays its role in promoting cooperation when the game is easy. We see from Table~\ref{tab:sn_metrics} that the real network has high diameter values and the clustering coefficient is 0. Therefore, the dynamics of the game with this network is equivalent to neither full assortative nor full disassortative networks (Table~\ref{tab:sn_metrics}). Instead, the low density and large diameter of the real network explain the slow decay of cooperation for large $\alpha$ values.

\subsection{\label{sec:results_probs_high_low}Balancing between the Subjective Audit Probability for High and Low Transactions} 

\begin{figure}
\centering

\includegraphics[width=0.46\textwidth]{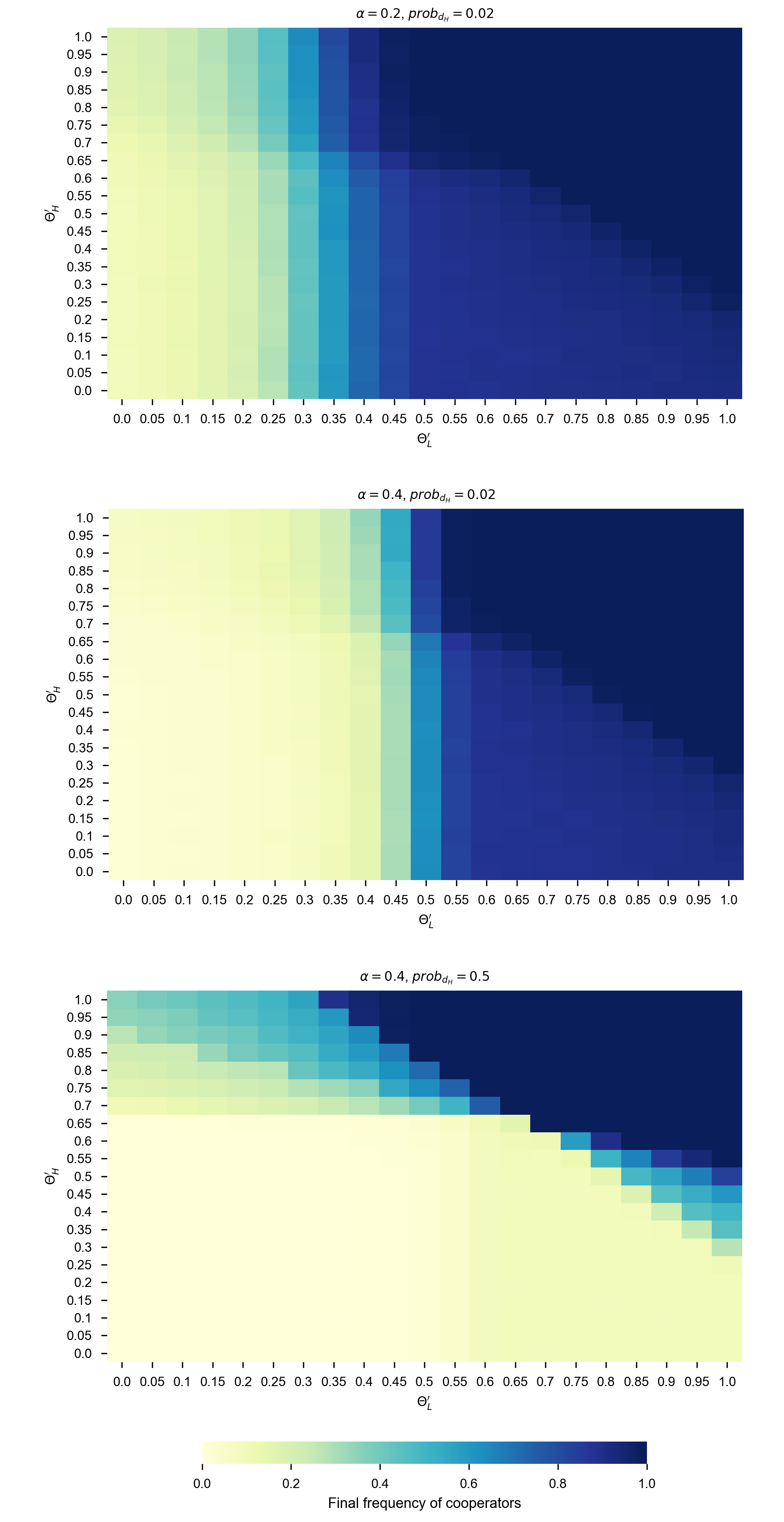}

\caption{The upper and middle heatmaps show sensitivity analysis on $\Theta'_L$ and $\Theta'_H$ for $\alpha=0.2$ and $\alpha=0.4$ (real data scenario where $prob_{d_H}=0.02$). The bottom heatmap shows the sensitivity analysis when $prob_{d_H}$ is $0.5$ and $\alpha=0.4$ for comparison. We observe that increasing the inspection probability for low transactions is preferable in the real world scenario where $prob_{d_H}=0.02$, but this conclusion does not apply when we have the same number of low and high transactions in the network (see the lower heatmap).}
\label{fig:different_prob_values}
\end{figure}

One of the main insights the analysis of real data from the Canarian tax agency revealed was the distinction between two types of transaction volumes: high and low. We would like to find the best policy to promote cooperation and correct tax paying behavior by determining the type of transaction the tax agency must focus on. In order to understand the impact of investigating these types of transactions, we use the evolutionary model to balance the focus on the subjective audit probability---which can be modulated differently depending on the transaction volume. Thus, we set different values for $\Theta'_H$ and $\Theta'_L$, which changes the construction of the probability linear function. We started from base values of $\Theta'_H = 0.5$ and $\Theta'_L = 0.5$ and considered a wide range of pairs for analysis, from 0 to 1 for both parameters. Figure~\ref{fig:different_prob_values} has three heatmaps showing the final frequency of cooperators for different subjective audit probabilities. The upper and middle plots show the results when $\alpha$ is equal to $0.2$ and $0.4$, respectively. The lower plot shows the dynamics when the numbers of high and low transactions are equal (i.e., $prob_{d_H}$ is $0.5$) and $\alpha=0.4$.

The analysis reflects important variations when modifying these subjective probabilities to favor a particular transaction type. We can see how tax fraud is limited when the subjective probability is higher for low transactions than high transactions. In fact, differences are not relevant when $\Theta'_L > 0.5$. However, when $\Theta'_L$ decreases, the number of cooperators declines almost independently of $\Theta'_H$. The number of cooperators declines and differences are significant when $\alpha$ values differ (e.g., cooperators are the dominant strategy in the final population only when $\alpha =0.05$ and the game is easy). The results change when the probability of high and low transactions is equal (Figure 9, lower plot). In this case, there are no major differences in the frequency of cooperators when $\Theta'_L$ and $\Theta'_H$ change. These results show that the significant effect of subjective probability on tax fraud, for low transactions in the upper and middle heatmaps, is mainly due to the larger number of low transactions in the network.

\subsection{\label{sec:diversity_pop}Population Diversity in the Subjective Inspection Probabilities} 

\begin{figure}
\centering
\includegraphics[width=0.45\textwidth]{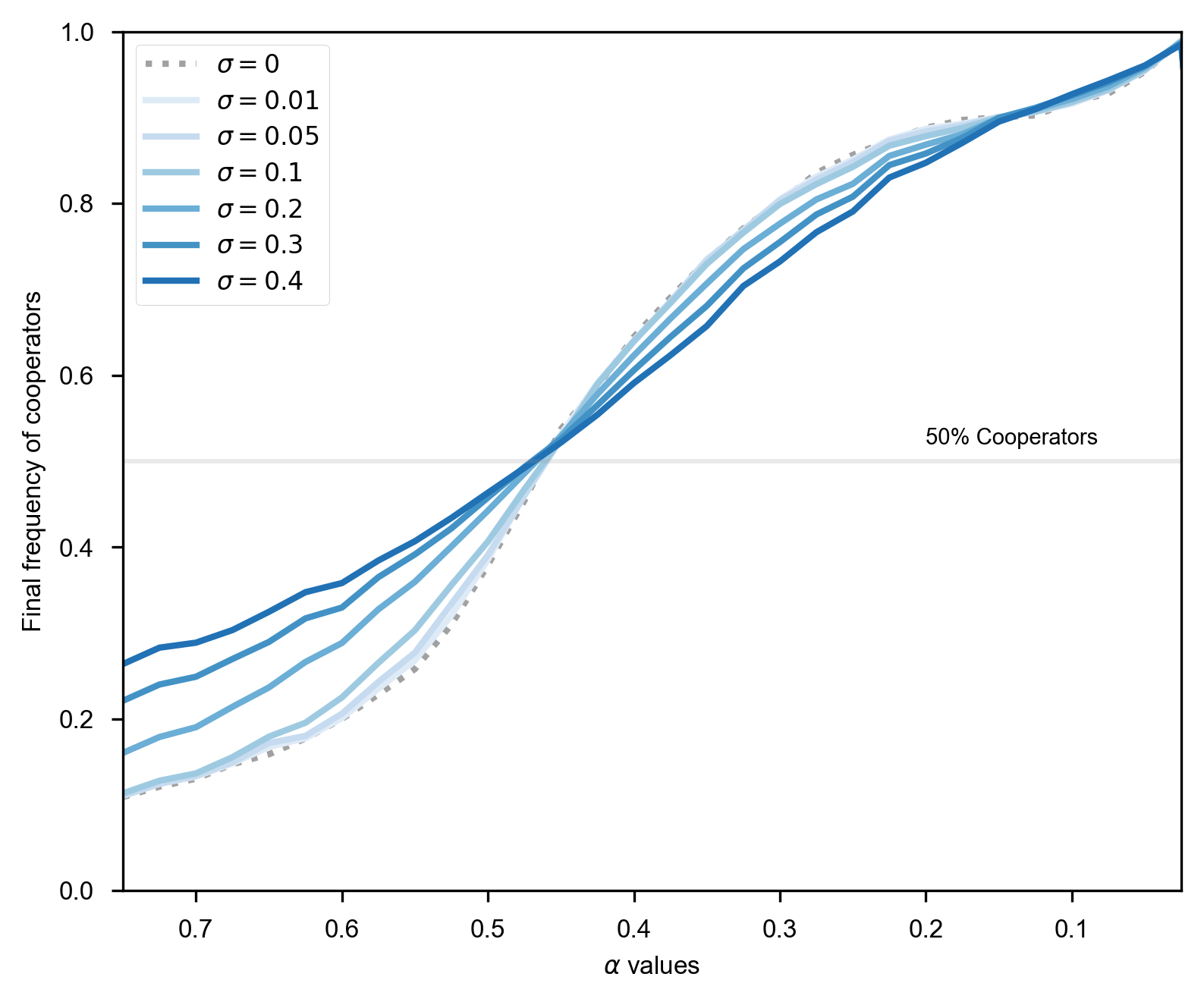}

\caption{Analysis of the diversity of subjective probabilities $\Theta'_L$ and $\Theta'_H$ when setting individuals of the population by generating a normal distribution $N(\mu, \sigma)$. We set $\mu$ as $0.5$ and plot different values of $\sigma$.}
\label{fig:stdev_subj_probs}
\end{figure}

Next, we analyze how diversity in the individuals of the population, with respect to their subjective probabilities, affects cooperation. In order to run this, we considered $\Theta'_H = \Theta'_L = 0.5$ as the mean $\mu$ of the normal distribution $N(\mu, \sigma)$ of the subjective probabilities of the whole population, and we modified the standard deviation $\sigma$ of the distribution. Figure~\ref{fig:stdev_subj_probs} shows the output of seven simulations with different standard deviation $\sigma$ values: from $0$, corresponding to the default configuration of the experiments in this work, to $0.4$, where individuals are highly diverse.

Figure~\ref{fig:stdev_subj_probs} shows how population diversity is beneficial for promoting cooperation when the game is hard (high values of $\alpha$), but cannot promote cooperation when the game is easy. Similar trends were observed when changing the density and other properties of the networks in Section~\ref{sec:SF_study}. This diversity changes the cooperation levels because of the polarization of the entire population, as observed by Antonioni et al.~\cite{Antonioni19}. Figure~\ref{fig:stdev_subj_probs} also reveals that diversity always induces a shift of the population to a 50\% polarization (gray horizontal line).

\subsection{\label{sec:results_rewards_fines}Impact of Rewarding and Penalty Policies} 

\begin{figure*}
\centering

 \includegraphics[width=0.32\textwidth]{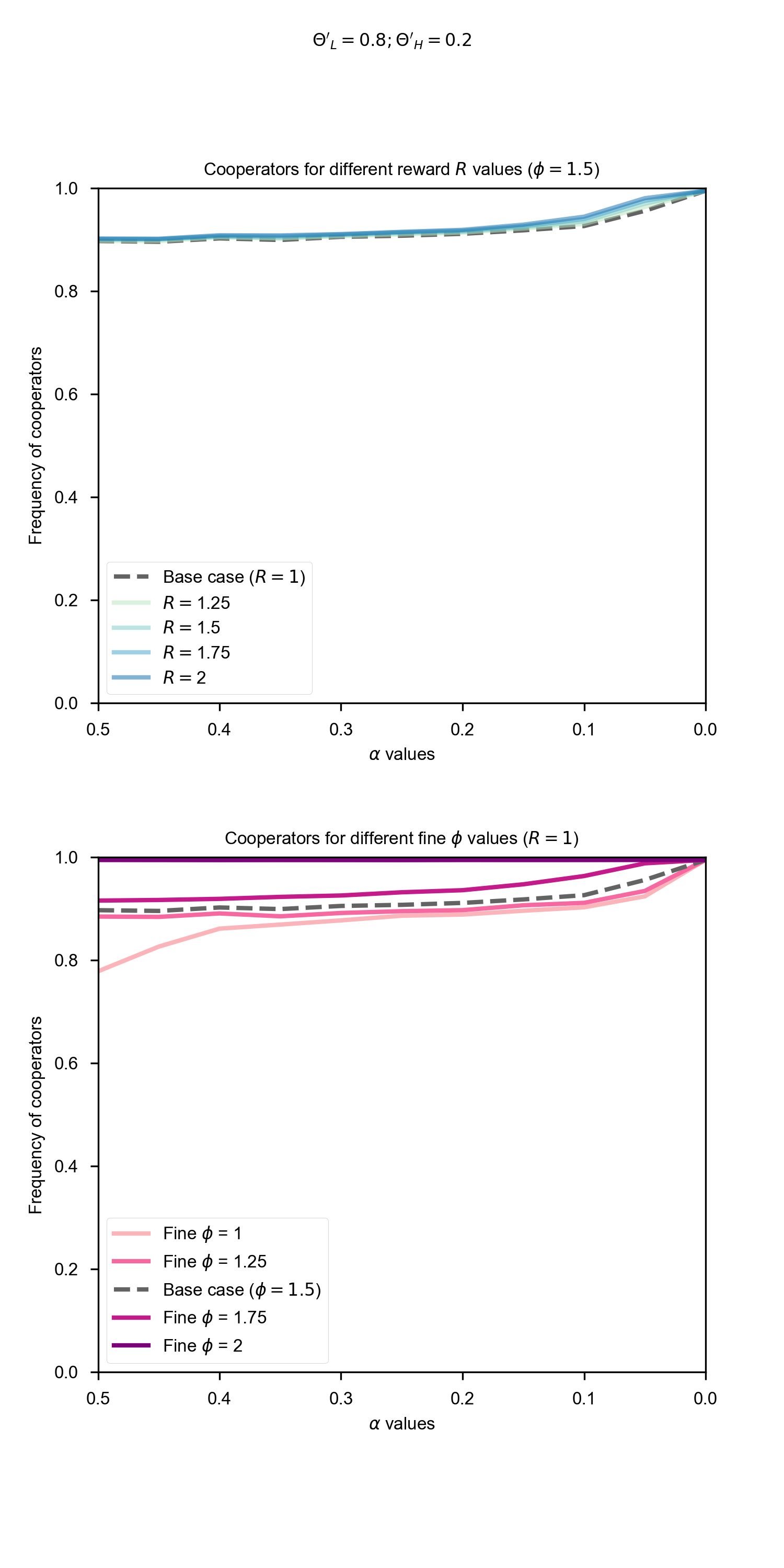}
 \includegraphics[width=0.32\textwidth]{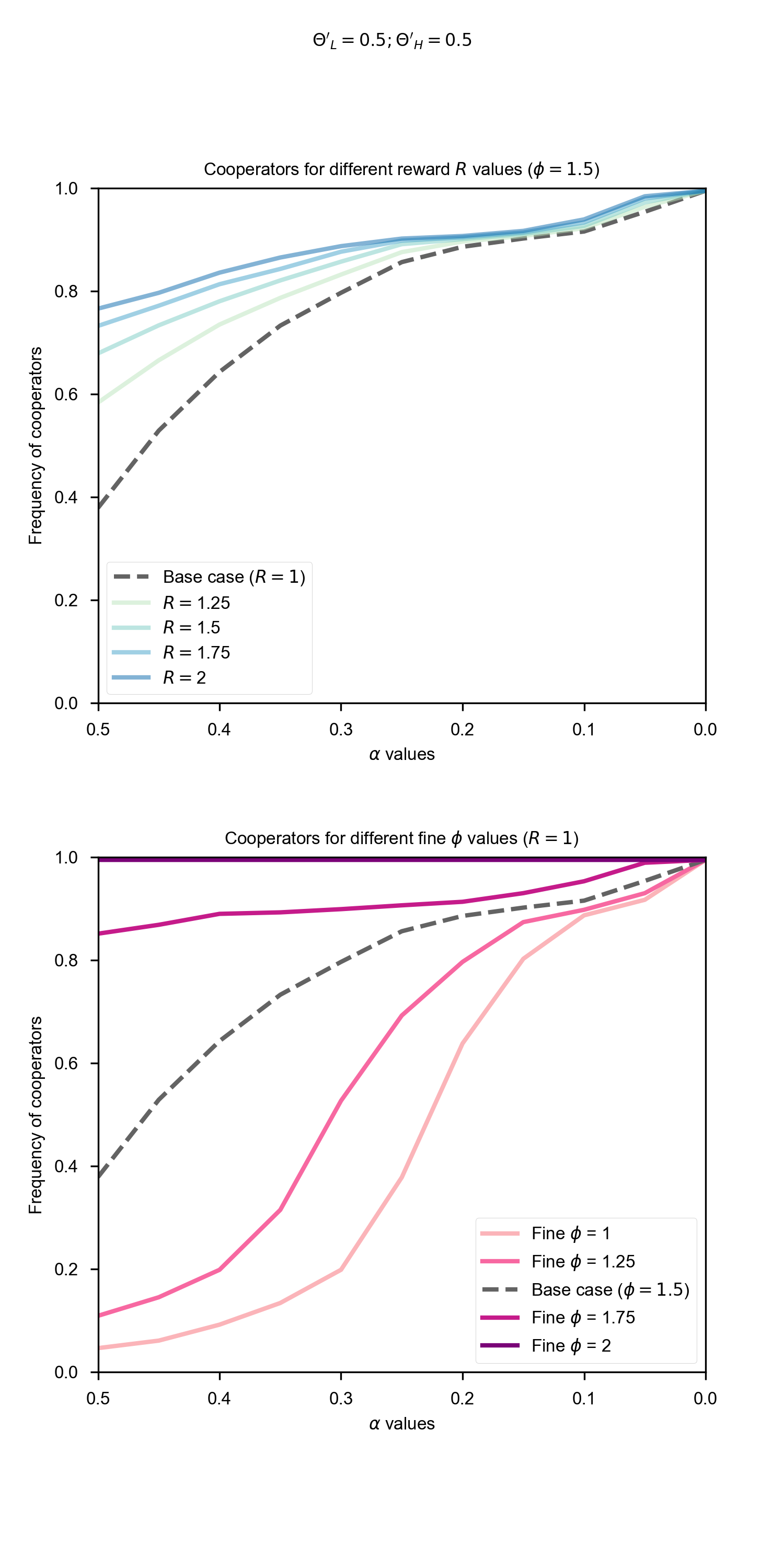}
 \includegraphics[width=0.32\textwidth]{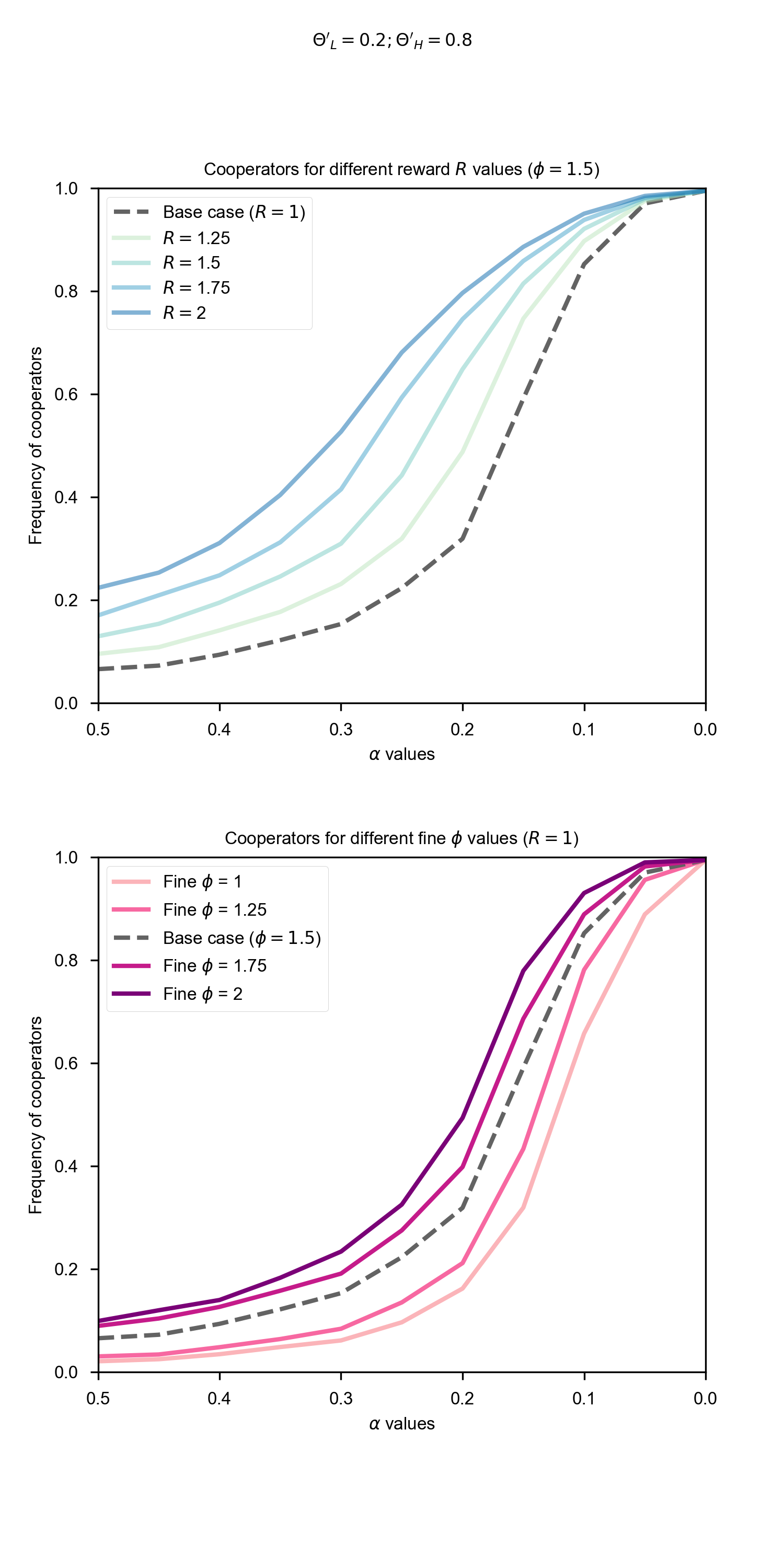}
 
\caption{Analysis of the reward $R$ (upper plots) and fine $\phi$ (bottom plots) parameters based on a sensitivity analysis on $\alpha$ for three subjective audit probability scenarios ($\Theta'_L = 0.8, \Theta'_H = 0.2$ in the first column, $\Theta'_L = \Theta'_H = 0.5$ in the second column, and $\Theta'_L =0.2, \Theta'_H = 0.8$ in the third column). We can see how the impacts of reward and punishment strategies depend on the scenario.}
\label{fig:SA_alphas_for_reward_fines}
\end{figure*}

In this final section of our model analysis, we focus on ascertaining if policies to increase the reward for cooperators are more efficient than those to increase the punishment for defectors via the fine values. Punishment versus reward has been studied in different public goods games and common pool resources~\cite{Fehr07,Chen15, Gois19}. For our analysis, we increased the values of reward $R$ from $1$ to $2$ and fines from $1$ (most liberal---defectors just have to return the unpaid tax) to $2$ (the fine is double of the unpaid quantity). Figure~\ref{fig:SA_alphas_for_reward_fines} shows the impact of different reward and fine values on cooperation under a sensitivity analysis of $\alpha$ and for three different scenarios of subjective audit probability $\Theta'_L$ and $\Theta'_H$.

We first observe how the impacts of both reward and fine policies differ depending on the subjective audit probability. When high transactions have a higher subjective audit  probability (the third scenario), increasing the reward, $R$, is more efficient for promoting cooperation than increasing the fine, $\phi$. The highest cooperation percentage is obtained with $R=2$ while keeping the base fine of $1.5$. However, for all the $\alpha$ values, increasing the fine up to $2$ does not generally induce as many new cooperators as rewarding policies do. 

When the subjective audit probability for low transactions is higher than for high transactions, the output of the model changes dramatically. If we are in the balanced second scenario ($\Theta'_H =0.5$ and $\Theta'_L = 0.5$), penalizing defectors with high fines is the most convenient option for promoting cooperation. In the first scenario, rewarding cooperators is almost invariant for the model dynamics. Therefore, both reward and punishment strategies must be carried out together while balancing the focus on either high or low transactions or, at least, apply them depending on the current scenario.

\section{\label{sec:conclusions}Final discussion}

We presented the first evolutionary game model for consumption taxes. This model represents cooperators and defectors and includes parameters to penalize tax evaders. It also considers the subjective probability of being inspected by the tax agency, which can be modulated with the size of the economic transaction. Players are linked through a scale-free network. Both the network topology and most of the model's parameters are fed with real data from the Canaries tax agency.

The stability and robustness of the model were demonstrated by simulating the effects of the undeclared quantity ($\alpha$ parameter), initial distribution of the population, and convergence to a steady state. After illustrating the main dynamics, we evaluated the two main questions for the tax agencies. First, we explored whether the agencies must focus on high or low transactions. We found that it is better to increase pressure on low transactions rather than high transactions. This is mainly due to the larger number of low transactions in the network. But the level of this pressure on low transactions is irrelevant once it is higher than the pressure on high transactions. This result is in line with previous findings~\cite{Fedeli1999} which support policies to increase audit probability but extend them by differentiating the audit probability according to the transaction size. Our results could encourage tax agencies to apply appropriate media actions targeting small transactions rather than high transactions. 

Second, our analysis showed that policies for either rewarding cooperators or punishing defectors must be executed in conjunction with policies for high and low transactions. For instance, we observed that, when the perceived inspection probability is more significant for high transactions, rewarding cooperators is more beneficial than increasing fines for defectors. This effect was consistent for different difficulty levels of the game (defined through the $\alpha$ parameter). However, when pressure is more important for low transactions, punishing defectors prevails as the best strategy. Our findings recommend tax agencies to follow a constructive approach to better reward companies behaving well by publicizing reward actions. These policies must be run in conjunction with other measures (e.g., balancing pressure on either low or high-value transactions).

The presented study has some limitations. The rewarding versus punishing policies do not take into account possible costs. A method of evaluating these two options by considering the costs for the agencies could be valuable. Researchers can also evaluate the response of temporal changes in the network topology. For example, one can study how temporal changes on the type of transaction between players (i.e., $d_{ij}$) can influence the game output as the employed payoff matrix would also change overtime. A more comprehensive study about diversity in the subjective inspection probabilities of the individuals can be performed too. Finally, a CI algorithm could identify the most influential companies (nodes) to be targeted with specific policies such as in Robles et al.~\cite{Robles20ESWA}.


 \section*{Acknowledgments} 
This work is jointly supported by the Spanish Ministry of Science, Andalusian Government, and ERDF under grants EXASOCO (PGC2018-101216-B-I00), SIMARK (P18-TP-4475) and RYC-2016-19800. The data used here is based on the project "Elaboration of a Computable General Equilibrium Model for the Analysis of Fiscal Policies on Sales Taxes" commissioned by the Economy and Finance Council of the Regional Government of the Canary Islands (FULP, CN45/08 240/57/100). We thank the Council of Economy, Knowledge and Employment of the Regional Government of the Canary Islands for allowing us to use the results of the latter project.
 

\bibliographystyle{IEEEtran}


\begin{thebibliography}{10}

\bibitem{Prichard2014}
W.~Prichard, A.~Cobham, and A.~Goodall, ``The ictd government revenue
  dataset,'' \emph{SSRN Electronic Journal}, no. September, 2014.

\bibitem{Keen2007}
M.~Keen, ``{VAT attacks !}'' \emph{International Tax and Public Finance},
  vol.~14, pp. 365--381, 2007.

\bibitem{DasGupta03}
A.~Das-Gupta and I.~Gang, ``Value added tax evasion, auditing and transactions
  matching,'' \emph{Institutional Elements of Tax Design}, pp. 25--48, 2003.

\bibitem{Nowak92}
M.~A. Nowak and R.~M. May, ``Evolutionary games and spatial chaos,''
  \emph{Nature}, vol. 359, no. 6398, pp. 826--829, 1992.

\bibitem{Shafi17}
K.~Shafi and H.~A. Abbass, ``A survey of learning classifier systems in
  games,'' \emph{IEEE Computational Intelligence Magazine}, vol.~12, no.~1, pp.
  42--55, 2017.

\bibitem{Vilone14}
D.~Vilone, J.~J. Ramasco, A.~S{\'a}nchez, and M.~San~Miguel, ``Social imitation
  versus strategic choice, or consensus versus cooperation, in the networked
  prisoner's dilemma,'' \emph{Physical Review E}, vol.~90, no.~2, p. 022810,
  2014.

\bibitem{Chiong12TEC}
R.~Chiong and M.~Kirley, ``Effects of iterated interactions in multiplayer
  spatial evolutionary games,'' \emph{IEEE Transactions on Evolutionary
  Computation}, vol.~16, no.~4, pp. 537--555, 2012.

\bibitem{Mittal09}
S.~Mittal and K.~Deb, ``Optimal strategies of the iterated prisoner's dilemma
  problem for multiple conflicting objectives,'' \emph{IEEE Transactions on
  Evolutionary Computation}, vol.~13, no.~3, pp. 554--565, 2009.

\bibitem{Hauert04}
C.~Hauert and M.~Doebeli, ``Spatial structure often inhibits the evolution of
  cooperation in the snowdrift game,'' \emph{Nature}, vol. 428, no. 6983, pp.
  643--646, 2004.

\bibitem{Wang06}
W.-X. Wang, J.~Ren, G.~Chen, and B.-H. Wang, ``Memory-based snowdrift game on
  networks,'' \emph{Physical Review E}, vol.~74, no.~5, p. 056113, 2006.

\bibitem{Chica18IEEETEC}
M.~Chica, R.~Chiong, M.~Kirley, and H.~Ishibuchi, ``A networked {N}-player
  trust game and its evolutionary dynamics,'' \emph{IEEE Transactions on
  Evolutionary Computation}, vol.~22, no.~6, pp. 866--878, 2018.

\bibitem{Chica19CNSNS}
M.~Chica, R.~Chiong, J.~J. Ramasco, and H.~Abbass, ``Effects of update rules on
  networked n-player trust game dynamics,'' \emph{Communications in Nonlinear
  Science and Numerical Simulation}, p. 104870, 2019.

\bibitem{Dawid08}
H.~Dawid, H.~La~Poutre, and X.~Yao, ``Computational intelligence in economic
  games and policy design,'' \emph{IEEE Computational Intelligence Magazine},
  vol.~3, no.~4, pp. 22--26, 2008.

\bibitem{Pickhardt14}
M.~Pickhardt and A.~Prinz, ``Behavioral dynamics of tax evasion--a survey,''
  \emph{Journal of Economic Psychology}, vol.~40, pp. 1--19, 2014.

\bibitem{Antoci14}
A.~Antoci, P.~Russu, and L.~Zarri, ``Tax evasion in a behaviorally
  heterogeneous society: An evolutionary analysis,'' \emph{Economic Modelling},
  vol.~42, pp. 106--115, 2014.

\bibitem{diMauro19}
L.~S. Di~Mauro, A.~Pluchino, and A.~E. Biondo, ``Tax evasion as a contagion
  game: evidences from an agent-based model,'' \emph{The European Physical
  Journal B}, vol.~92, no.~5, p. 103, 2019.

\bibitem{degiovanni19}
D.~De~Giovanni, F.~Lamantia, and M.~Pezzino, ``A behavioral model of
  evolutionary dynamics and optimal regulation of tax evasion,''
  \emph{Structural Change and Economic Dynamics}, vol.~50, pp. 79--89, 2019.

\bibitem{Macal05}
C.~M. Macal and M.~J. North, ``Tutorial on agent-based modeling and
  simulation,'' in \emph{Proceedings of the 37th conference on Winter
  simulation}.\hskip 1em plus 0.5em minus 0.4em\relax ACM, 2005, pp. 2--15.

\bibitem{Chica17JMR}
M.~Chica and W.~Rand, ``Building agent-based decision support systems for
  word-of-mouth programs. a freemium application,'' \emph{Journal of Marketing
  Research}, vol.~54, pp. 752--767, October 2017.

\bibitem{Abbass11}
H.~Abbass, A.~Bender, S.~Gaidow, and P.~Whitbread, ``Computational red teaming:
  Past, present and future,'' \emph{IEEE Computational Intelligence Magazine},
  vol.~6, no.~1, pp. 30--42, 2011.

\bibitem{Newman06}
M.~Newman, A.-L. Barab{\'a}si, and D.~J. Watts, \emph{The structure and
  dynamics of networks}.\hskip 1em plus 0.5em minus 0.4em\relax Princeton
  University Press, 2006.

\bibitem{Santos08}
F.~C. Santos, M.~D. Santos, and J.~M. Pacheco, ``Social diversity promotes the
  emergence of cooperation in public goods games,'' \emph{Nature}, vol. 454,
  no. 7201, pp. 213--216, 2008.

\bibitem{Amaral16}
M.~A. Amaral, L.~Wardil, M.~Perc, and J.~K. da~Silva, ``Evolutionary mixed
  games in structured populations: Cooperation and the benefits of
  heterogeneity,'' \emph{Physical Review E}, vol.~93, no.~4, p. 042304, 2016.

\bibitem{Broido19}
A.~D. Broido and A.~Clauset, ``Scale-free networks are rare,'' \emph{Nature
  communications}, vol.~10, no.~1, pp. 1--10, 2019.

\bibitem{Allingham1972}
M.~G. Allingham and A.~Sandmo, ``{Income Tax Evasion: A Theoretical
  Analysis},'' \emph{Journal of Public Economics}, vol.~1, pp. 323--338, 1972.

\bibitem{Bordignon1993}
M.~Bordignon, ``{A fairness approach to lncome tax},'' \emph{Journal of Public
  Economics}, vol.~52, pp. 345--362, 1993.

\bibitem{Bazart2014}
C.~Bazart and A.~Bonein, ``{Reciprocal relationships in tax compliance
  decisions},'' \emph{Journal of Economic Psychology}, vol.~40, pp. 83--102,
  2014.

\bibitem{Traxler2010}
C.~Traxler, ``{European Journal of Political Economy Social norms and
  conditional cooperative taxpayers},'' \emph{European Journal of Political
  Economy}, vol.~26, no.~1, pp. 89--103, 2010.

\bibitem{Alm2012}
J.~Alm, ``{Measuring, explaining, and controlling tax evasion: lessons from
  theory , experiments, and field studies},'' \emph{International Tax and
  Public Finance}, vol.~19, pp. 54--77, 2012.

\bibitem{Alm2015b}
J.~Alm, K.~M. Bloomquist, and M.~McKee, ``{On the external validity of
  laboratory tax compliance experiments},'' \emph{Economic Inquiry}, vol.~53,
  no.~2, pp. 1170--1186, 2015.

\bibitem{Bonein2018}
A.~Bonein, ``{La r{\'{e}}ciprocit{\'{e}} , entre psychologie et
  rationalit{\'{e}} {\'{e}}conomique},'' \emph{Revue Fran{\c{c}}aise
  d'{\'{e}}conomie}, vol.~23, pp. 203--232, 2018.

\bibitem{Gintis2000}
H.~Gintis, ``{Strong Reciprocity and Human Sociality},'' \emph{Journal of
  Theoretical Biiology}, vol. 206, pp. 169--179, 2000.

\bibitem{Frey2007}
B.~S. Frey and B.~Torgler, ``{Tax morale and conditional cooperation},''
  \emph{Journal of Comparative Economics}, vol.~35, pp. 136--159, 2007.

\bibitem{Blaufus2017}
K.~Blaufus, J.~Bob, P.~E. Otto, and N.~Wolf, ``{The Effect of Tax Privacy on
  Tax Compliance – An Experimental Investigation The Effect of Tax Privacy on
  Tax Compliance – An Experimental Investigation},'' \emph{European
  Accounting Review}, vol.~26, no.~3, pp. 561--580, 2017.

\bibitem{Calvet2014}
R.~Calvet and J.~Alm, ``{Empathy , sympathy , and tax compliance},''
  \emph{Journal of Economic Psychology}, vol.~40, pp. 62--82, 2014.

\bibitem{Fedeli1999}
S.~Fedeli and F.~Forte, ``{Joint income-tax and VAT-chain evasion},''
  \emph{European Journal of Political Economy}, vol.~15, pp. 391--415, 1999.

\bibitem{Abraham2017}
M.~Abraham, K.~Lorek, F.~Richter, and M.~Wrede, ``{Collusive tax evasion and
  social norms},'' \emph{International Tax and Public Finance}, vol.~24, no.~2,
  pp. 179--197, 2017.

\bibitem{Boadway2002}
R.~Boadway, N.~Marceau, and S.~Mongrain, ``{Joint tax evasion},''
  \emph{Canadian Journal of Economics}, vol.~35, no.~3, pp. 417--435, 2002.

\bibitem{Bloomquist2006}
K.~M. Bloomquist, ``A comparison of agent-based models of income tax evasion,''
  \emph{Science Computer Review}, vol.~24, no.~4, pp. 411--425, 2006.

\bibitem{Hokamp2018}
S.~Hokamp, L.~Gulyas, M.~Koehler, and S.~Wijesinghe, \emph{Agent-based Modeling
  of Tax Evasion}.\hskip 1em plus 0.5em minus 0.4em\relax Wiley, 2018.

\bibitem{Chen08}
S.-H. Chen, ``Software-agent designs in economics: An interdisciplinary,''
  \emph{IEEE Computational Intelligence Magazine}, vol.~3, no.~4, pp. 18--22,
  2008.

\bibitem{Korobow2007}
C.~Korobow, A.and~Johnson and R.~Axtell, ``An agent-based model of tax
  compliance with social networks,'' \emph{National Tax Journal}, vol.~60,
  no.~3, pp. 411--425, 2007.

\bibitem{Hashimzade2014}
N.~Hashimzade, G.~Myles, F.~Page, and M.~Rablen, ``Social networks and
  occupational choice: The endogenous formation of attitudes and beliefs about
  tax compliance,'' \emph{Journal of Economic Psychology}, vol.~40, pp.
  134--146, 2014.

\bibitem{Hashimzade2015}
N.~Hashimzade, G.~D. Myles, F.~Page, and M.~Rablen, ``The use of agent-based
  modelling to investigate tax compliance,'' \emph{Economics of Governance},
  vol.~16, no.~2, pp. 143--164, 2015.

\bibitem{Hashimzade2017}
N.~Hashimzade and G.~Myles, ``Risk-based audits in a behavioral model,''
  \emph{Public Finance Review}, vol.~45, no.~1, pp. 140--165, 2017.

\bibitem{Bloomquist2011}
K.~Bloomquist, ``Tax compliance as an evolutionary coordination game: An
  agent-based approach,'' \emph{Public Finance Review}, vol.~39, no.~1, pp.
  25--49, 2011.

\bibitem{PickhardtSeibold2014}
M.~Pickhardt and G.~Seibold, ``Income tax evasion dynamics: Evidence from an
  agent-based econophysics model,'' \emph{Journal of Economic Psychology},
  vol.~40, pp. 147--160, 2014.

\bibitem{Andrei2014}
A.~L. Andrei, K.~Comer, and M.~Koehler, ``An agent-based model of network
  effects on tax compliance and evasion,'' \emph{Journal of Economic
  Psychology}, vol.~40, pp. 119--133, 2014.

\bibitem{Bloomquist2012}
K.~M. Bloomquist, ``Incorporating indirect effects in audit case selection: An
  agent-based approach,'' \emph{IRS document}, pp. 1--14, 2012.

\bibitem{Wardil13}
L.~Wardil and J.~K. da~Silva, ``The evolution of cooperation in mixed games,''
  \emph{Chaos, Solitons \& Fractals}, vol.~56, pp. 160--165, 2013.

\bibitem{Allen2015}
B.~Allen and M.~A. Nowak, ``Games among relatives revisited,'' \emph{Journal of
  theoretical biology}, vol. 378, pp. 103--116, 2015.

\bibitem{Nowak10}
M.~A. Nowak, C.~E. Tarnita, and T.~Antal, ``Evolutionary dynamics in structured
  populations,'' \emph{Philosophical Transactions of the Royal Society of
  London B: Biological Sciences}, vol. 365, no. 1537, pp. 19--30, 2010.

\bibitem{Gois19}
A.~R. G{\'o}is, F.~P. Santos, J.~M. Pacheco, and F.~C. Santos, ``Reward and
  punishment in climate change dilemmas,'' \emph{Scientific reports}, vol.~9,
  no.~1, pp. 1--9, 2019.

\bibitem{Vasconcelos14}
V.~V. Vasconcelos, F.~C. Santos, J.~M. Pacheco, and S.~A. Levin, ``Climate
  policies under wealth inequality,'' \emph{Proceedings of the National Academy
  of Sciences}, vol. 111, no.~6, pp. 2212--2216, 2014.

\bibitem{Clauset2009a}
A.~Clauset, C.~R. Shalizi, and M.~E. Newman, ``{Power-law distributions in
  empirical data},'' \emph{SIAM Review}, vol.~51, no.~4, pp. 661--703, 2009.

\bibitem{PYME19}
IPYME, ``Estadísticas pyme: Evolución e indicadores,'' www.ipyme.org,
  Gobierno de España, March 2019.

\bibitem{Barabasi99}
A.~L. Barab{\'a}si and R.~Albert, ``Emergence of scaling in random networks,''
  \emph{Science}, vol. 286, no. 5439, pp. 509--512, 1999.

\bibitem{Xulvi04}
R.~Xulvi-Brunet and I.~M. Sokolov, ``Reshuffling scale-free networks: From
  random to assortative,'' \emph{Physical Review E}, vol.~70, no.~6, p. 066102,
  2004.

\bibitem{Antonioni19}
A.~Antonioni, L.~A. Martinez-Vaquero, C.~Mathis, L.~Peel, and M.~Stella,
  ``Individual perception dynamics in drunk games,'' \emph{Physical Review E},
  vol.~99, no.~5, p. 052311, 2019.

\bibitem{Fehr07}
E.~Fehr and K.~M. Schmidt, ``Adding a stick to the carrot? the interaction of
  bonuses and fines,'' \emph{American Economic Review}, vol.~97, no.~2, pp.
  177--181, 2007.

\bibitem{Chen15}
X.~Chen, T.~Sasaki, {\AA}.~Br{\"a}nnstr{\"o}m, and U.~Dieckmann, ``First
  carrot, then stick: how the adaptive hybridization of incentives promotes
  cooperation,'' \emph{Journal of the royal society interface}, vol.~12, no.
  102, p. 20140935, 2015.

\bibitem{Robles20ESWA}
J.~F. Robles, M.~Chica, and O.~Cordon, ``Evolutionary multiobjective
  optimization to target social network influentials in viral marketing,''
  \emph{Expert Systems with Applications}, vol.~9, p. 113183, 2020.

\end{thebibliography}

\end{document}